# Outperforming Self-Attention Mechanisms in Solar Irradiance Forecasting via Physics-Guided Neural Networks

Mohammed Ezzaldin Babiker Abdullah[a], Rufaidah Abdallah Ibrahim Mohammed[a]
Omdurman Islamic University, Omdurman, Sudan.
Izzeldeenm@gmail.com, Rufaidahabdallah@gmail.com
Corresponding author: Izzeldeenm@gmail.com

**Abstract**

Accurate Global Horizontal Irradiance (GHI) forecasting is critical for grid stability, particularly in arid regions characterized by rapid aerosol fluctuations. While recent trends favor computationally expensive Transformer-based architectures, this paper challenges the prevailing "complexity-first" paradigm. We propose a lightweight, Physics-Informed Hybrid CNN-BiLSTM framework that prioritizes domain knowledge over architectural depth. The model integrates a Convolutional Neural Network (CNN) for spatial feature extraction with a Bi-Directional LSTM for capturing temporal dependencies. Unlike standard data-driven approaches, our model is explicitly guided by a vector of 15 engineered features—including Clear-Sky indices and Solar Zenith Angle—rather than relying solely on raw historical data. Hyperparameters are rigorously tuned using Bayesian Optimization to ensure global optimality. Experimental validation using NASA POWER data in Sudan demonstrates that our physics-guided approach achieves a Root Mean Square Error (RMSE) of 19.53 W/m², significantly outperforming complex attention-based baselines (RMSE 30.64 W/m²). These results confirm a "Complexity Paradox": in high-noise meteorological tasks, explicit physical constraints offer a more efficient and accurate alternative to self-attention mechanisms. The findings advocate for a shift towards hybrid, physics-aware AI for real-time renewable energy management.

## 1. Introduction

The global energy sector is witnessing a rapid trend towards integrating photovoltaic (PV) energy systems at high penetration levels to meet the growing demand for clean energy; however, this expansion imposes unprecedented operational challenges on the stability of electrical grids (Bank et al., 2013). This complexity stems primarily from the stochastic and volatile nature of solar irradiance, where sharp instantaneous changes in power, known as ramp-rate events, cause critical imbalances between generation and loads (Paletta et al., 2023). These operational dilemmas are exceptionally exacerbated in arid and semi-arid regions, such as Sudan, where challenges are not limited to transient cloud dynamics but extend to include severe impacts resulting from high aerosol loading and the frequency of intense dust storms (Kosmopoulos et al., 2017). These environmental factors play a crucial role in the sudden scattering and absorption of Global Horizontal Irradiance (GHI), creating highly complex optical and physical conditions that traditional forecasting models fail to track accurately.

Despite extensive research in the field of solar irradiance forecasting and model development, significant challenges persist in reliably generating accurate, short-term forecasts under these harsh climatic conditions. Traditional methods, such as physical parametric models and statistical approaches like ARIMA, have shown clear shortcomings in capturing deep non-linear dependencies during rapid irradiance fluctuations (Voyant et al., 2017). With the advent of deep learning, architectures based on Convolutional Neural Networks (CNN) and Long Short-Term Memory (LSTM) demonstrated improved capabilities in extracting spatial and temporal features; yet, they often operate as black boxes lacking the ability to generalize across different climatic zones without extensive retraining (Qing and Niu, 2018 Recently, research has shifted towards hybrid architectures and models relying on Self-Attention mechanisms and Transformers, operating under the assumption that increasing architectural complexity inherently leads to better forecasting accuracy (Al-Ali et al., 2023; Hou et al., 2023; Liu et al., 2023; Pospíchal et al., 2022). However, recent critical evaluations reveal that Transformers can severely degrade in performance during continuous time-series forecasting, as they often lose the precise temporal ordering of data, rendering them less effective than simpler, well-structured models (Zeng et al., 2023).

This over-reliance on the Complexity-First paradigm suffers from fundamental limitations, most notably the complete disregard for the explicit physical laws governing atmospheric dynamics. This omission frequently leads to critical issues such as Phase Lag and data over-smoothing, diminishing the operational reliability of models in dust-dense and highly volatile environments. Relying on complex attention mechanisms to implicitly infer these laws from raw data requires massive computational resources and fails to guarantee stable performance (Paletta et al., 2023). Conversely, Physics-Informed Machine Learning directly addresses this gap by integrating deterministic physical laws, such as clear-sky indices and solar geometry, to constrain neural networks, preventing physically implausible predictions and ensuring robust generalization in extreme climates (Karniadakis et al., 2021). Furthermore, adopting lightweight deep learning architectures combined with advanced hyperparameter tuning, such as Bayesian Optimization, drastically reduces computational overhead while maximizing real-time responsiveness (Hafeez et al., 2024). These strategic integrations highlight the urgent need for a physics-guided, computationally efficient framework in real-time forecasting, establishing it as a fundamental pillar for grid security and optimal Maximum Power Point Tracking (MPPT) control.

The domain of solar irradiance forecasting has witnessed a paradigm shift over the past decades, evolving from simple empirical correlations to sophisticated data-driven architectures. Early approaches relied heavily on physical parametric models, such as those proposed by Kasten and Czeplak, which utilize deterministic solar geometry and atmospheric turbidity to estimate clear-sky radiation. While physically interpretable, these models fail to account for stochastic cloud dynamics. Subsequently, statistical methods like ARIMA and SVR gained prominence, with foundational time-series methodologies established by Box et al. However, comprehensive reviews by Voyant et al. (2017) demonstrated that while statistical models perform well in stable weather, they exhibit significant limitations during rapid irradiance fluctuations due to their inability to capture deep non-linear dependencies.

With the advent of high-performance computing, deep learning became the standard for handling non-linear meteorological data. Hochreiter and Schmidhuber introduced the LSTM network, which solved the vanishing gradient problem in time-series data. More recently, Qing and Niu (2018) successfully applied LSTM to solar forecasting, outperforming traditional machine learning methods. Similarly, CNNs were adapted by Zang et al. to extract spatial features from weather data. However, as noted by Ghimire et al., purely data-driven single models often function as black boxes, struggling to generalize across different climatic zones without extensive retraining.

The current state-of-the-art focuses on hybridizing architectures to maximize feature extraction. Alharkan et al. proposed a dual-stream CNN-LSTM model, while Li et al. explored GANs for synthetic data augmentation. Most notably, the trend has shifted towards Transformers and Self-Attention mechanisms, as seen in the recent works of Al-Ali et al. (2023) and Hou et al. (2023), who argued that attention layers capture long-term dependencies better than recurrence. Furthermore, Radzi et al. (2025) and Herrera-Casanova et al. (2024) highlighted the importance of hyperparameter optimization in enhancing model stability.

Despite these advancements, a critical gap remains. The prevailing trend assumes that increasing architectural complexity inherently leads to better accuracy. This work challenges this Complexity-First paradigm. We hypothesize that for solar irradiance, which is governed by known physical laws, incorporating explicit physical knowledge is computationally more efficient and accurate than relying on the model to implicitly learn these laws through complex attention mechanisms. While deep learning models like LSTM and CNN have become the de facto standard for time-series forecasting, they fundamentally operate as black boxes. Purely data-driven approaches rely on identifying statistical patterns in historical data, often ignoring the governing physical laws of atmospheric dynamics. Consequently, these models suffer from critical limitations, most notably Phase Lag, and poor generalization in unseen weather conditions. Furthermore, the recent trend in literature has shifted towards increasingly complex architectures, often increasing computational cost without yielding proportional accuracy gains in relatively simple, low-dimensional weather tasks.

To address these limitations, this paper proposes a Physics-Informed Hybrid Framework that marries the pattern-recognition power of deep learning with the deterministic laws of solar geometry. Unlike standard models, the proposed system is guided by an engineered feature vector that injects explicit physical knowledge, such as clear-sky radiation, zenith angle, and atmospheric volatility, directly into the learning process. The primary contributions of this study are fourfold. First, a Physics-Informed Hybrid Architecture is proposed: a lightweight hybrid model (CNN-BiLSTM) that integrates spatial feature extraction with bidirectional temporal learning, explicitly guided by a vector of 15 engineered physical features to bridge the gap between physical laws and data-driven learning. Second, this study provides a validation of the Complexity Paradox, empirically demonstrating that increasing model complexity does not guarantee superior performance in high-noise environments, as our physics-guided model outperforms complex state-of-the-art Transformer and Attention-based architectures in arid climates. Third, Bayesian Hyperparameter Optimization is employed to identify the optimal hyperparameters for the CNN and BiLSTM layers, ensuring reproducibility and maximizing stability against local minima. Fourth, the model demonstrates robustness in arid and dusty climates through a specialized case study using NASA POWER data for Sudan, a region characterized by high aerosol optical depth and rapid irradiance fluctuations, proving the model's superior capability in handling stochastic phase lag errors compared to traditional statistical baselines.

The remainder of this paper is organized as follows. Section 2 details the methodology, including data acquisition, physics-based feature engineering, the proposed hybrid CNN-BiLSTM architecture, and the Bayesian optimization framework used for hyperparameter tuning. Section 3 presents a comprehensive analysis of the experimental results, comparing our approach against state-of-the-art baselines. Finally, Section 4 concludes the study and outlines future research directions.

## 2. Methodology

### 2.1 Experimental Data and Site Description

In this study, we propose a comprehensive Physics-Informed Deep Learning framework designed to address the stochastic challenges of solar irradiance forecasting in arid regions. As visually summarized in Fig. 1, the methodology follows a multi-stage pipeline: (1) Data Acquisition and Physics Injection, where raw meteorological inputs are augmented with deterministic solar laws; (2) The Hybrid Processing Core, which normalizes features and maps them through the proposed CNN-BiLSTM architecture; and (3) The Application Stage, delivering actionable GHI forecasts for grid management.

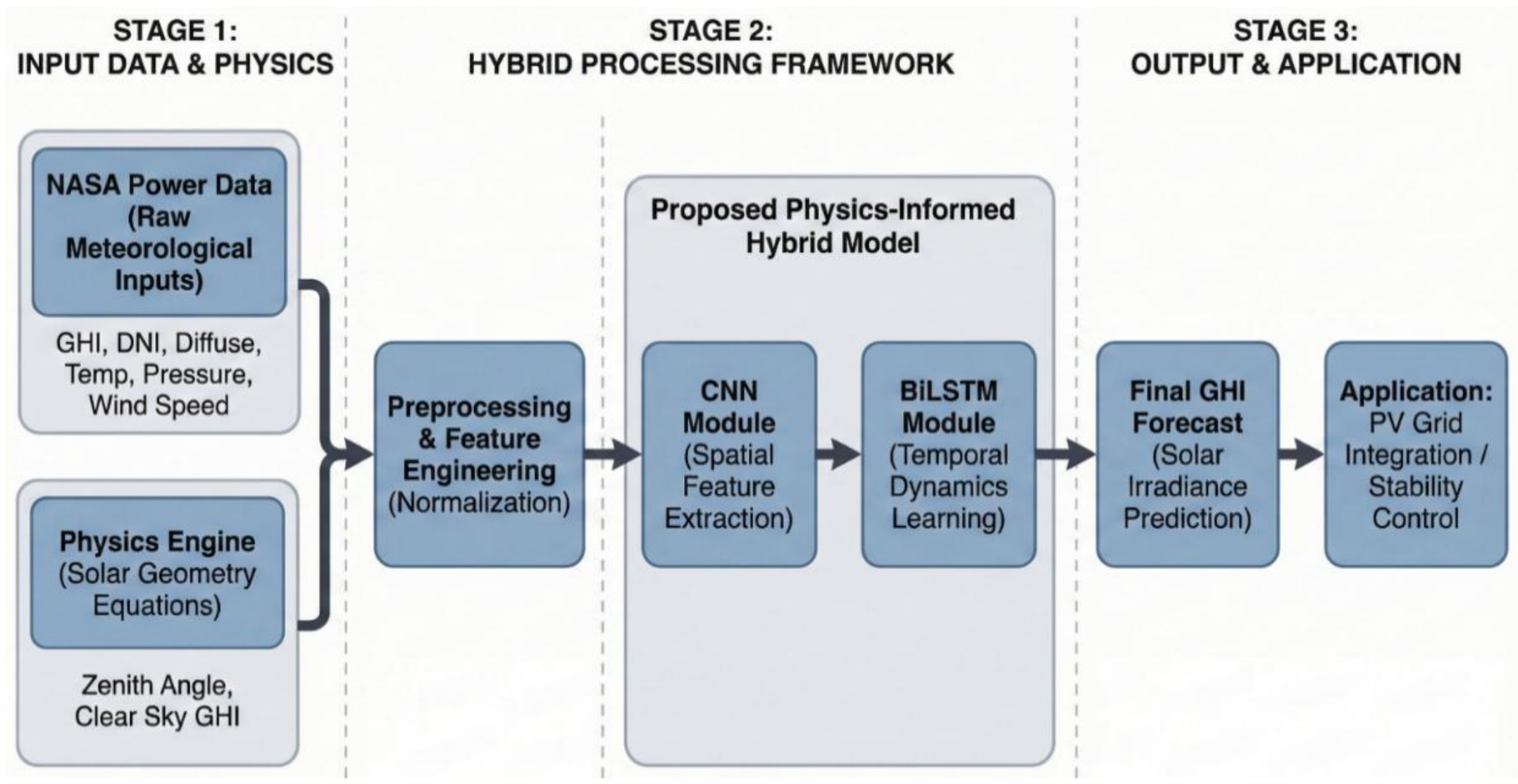

**Fig. 1.** High-level schematic of the proposed Physics-Informed Framework, illustrating the end-to-end workflow from data acquisition and physical feature engineering to the final predictive application.

1. Study Location and Data Source:

The experimental validation focuses on Omdurman, Sudan (14.7°N, 33.2°E), a region representative of semi-arid climates characterized by high solar potential and significant aerosol challenges. High-resolution meteorological data were acquired from the NASA POWER database, derived from satellite observations and global reanalysis models. The dataset consists of hourly records covering two distinct temporal phases to ensure rigorous evaluation.

2. Chronological Data Splitting Strategy:

To simulate a realistic forecasting scenario and prevent data leakage, a strict chronological splitting approach was adopted rather than random shuffling. The dataset is divided as follows:

- Model Development Phase (2010–2015): This period is partitioned into 70% for training, 15% for validation (used for Bayesian optimization and early stopping), and 15% for internal testing to evaluate performance on unseen data from the same epoch.
- Stress-Testing Phase (2020–2024): A completely independent dataset spanning five recent years was reserved for external testing. The 5-year gap between training and testing serves as a "temporal stress test" to verify that the model has learned the underlying physical laws rather than merely memorizing temporary weather patterns.

3. Input Feature Matrix:

The model input vector consists of 15 carefully selected features representing the thermodynamic and radiative state of the atmosphere. As detailed in Table 1, the primary inputs include Global Horizontal Irradiance (GHI), Direct Normal Irradiance (DNI),

Diffuse Irradiance (DHI), Relative Humidity (RH), Dew Point Temperature ($T_{\{dew\}}$), Wet Bulb Temperature ($T_{\{wet\}}$), and Ambient Temperature ($T_{\{amb\}}$). These observations are augmented with physics-derived variables (Calculated Clear-Sky GHI, Calculated Clearness Index $KT_{\{calc\}}$, Satellite Clearness Index $KT_{\{sat\}}$, and Volatility Index) and cyclical time encodings (Sin/Cos of Hour and Day) to capture diurnal and seasonal periodicities.

**Table 1**. Description of physical and meteorological input features

| Feature Category | Variable | Symbol | Unit | Physical Role & Description |
|---|---|---|---|---|
| Target | Global Horizontal Irradiance | $GHI$ | $W/m^2$ | Primary target variable representing total solar radiation received on a horizontal surface. |
| Physics-Derived | Clear-Sky GHI | $GHI_{cs}$ | $W/m^2$ | Theoretical baseline irradiance calculated locally to filter out geometric effects. |
| Physics-Derived | Calculated Clearness Index | $KT_{calc}$ | | Ratio of GHI to Clear-Sky GHI; indicates instantaneous atmospheric transmissivity (Linear scale). |
| Physics-Derived | Satellite Clearness Index | $KT_{sat}$ | | Satellite-derived cloudiness measure obtained from NASA data (Log-transformed). |
| Physics-Derived | Volatility Index | $Vol$ | | Rolling standard deviation of $KT_{calc}$; measures weather stability and rapid fluctuations. |
| Solar Component | Direct Normal Irradiance | $DNI$ | $W/m^2$ | Direct component of radiation; critical for determining cloud density and type. |
| Solar Component | Diffuse Horizontal Irradiance | $DHI$ | $W/m^2$ | Scattered radiation component; reflects the extent of cloud cover and atmospheric scattering. |
| Meteorological | Relative Humidity | $RH$ | | Atmospheric moisture content; strongly correlated with light scattering and cloud formation. |
| Meteorological | Dew Point Temperature | $T_{dew}$ | $C^\circ$ | The temperature at which air becomes saturated; a precise indicator of atmospheric water vapor. |
| Meteorological | Ambient Temperature | $T_{amb}$ | $C^\circ$ | Surface air temperature; influences PV cell efficiency (Log-transformed input). |
| Meteorological | Wet Bulb Temperature | $T_{wet}$ | $C^\circ$ | Thermodynamic temperature indicating evaporative cooling potential. |
| Temporal | Raw Timestamp | $Y, M, D, H$ | | Discrete raw time features (Year, Month, Day, Hour) used to identify exact chronological position without encoding. |

## 2.2 Data Preprocessing and Physical Consistency Checks

Raw satellite and reanalysis data inevitably contain artifacts such as sensor noise, missing values, or non-physical outliers. To ensure the integrity of the training process, a rigorous multi-stage preprocessing protocol was implemented:

1. Statistical Cleaning and Imputation:

The dataset was scanned for physically impossible values (e.g., negative irradiance or wind speed) and standard error codes (e.g., -999). These anomalies were treated as missing data. To preserve the continuity of the time series without discarding valuable records, small gaps were bridged using linear interpolation. This method was selected due to the naturally smooth transitions of meteorological variables at hourly resolutions, ensuring minimal distortion of the underlying trends.

2. Temporal Alignment:

Given that BiLSTM networks require strictly sequential input, the timeline was standardized to a constant 1-hour resolution ($\Delta t = 3600s$). Duplicate timestamps were purged, and temporal discontinuities were padded to ensure a continuous, equidistant time series, preventing the model from learning incorrect temporal dependencies.

3. Physics-Based Night Masking: To eliminate digital noise often present in satellite products during low-irradiance periods, a physics-guided "Night Mask" was applied. Nighttime was dynamically defined as intervals where the locally calculated theoretical Clear-Sky GHI fell below a threshold of 0 W/m². During these intervals, all radiative variables (e.g., GHI) were forcibly clamped to zero. This step constrains the optimization process, preventing the network from learning spurious correlations from night-time sensor noise.

4. Hybrid Normalization Strategy: Statistical analysis revealed significant disparity in feature scales (e.g., ambient temperature in C° vs. clearness index in fractions). To address this, a hybrid normalization approach was adopted. First, the input features were standardized using Z-Score normalization (zero mean, unit variance) to handle outliers and facilitate gradient convergence. Second, the target variable (GHI) was scaled using Min-Max normalization to align with the sensitive range of the output activation function, ensuring numerical stability.

## 2.3. Physics-Guided Feature Engineering

Instead of relying solely on the neural network to implicitly learn the laws of physics from massive datasets, this study employs a "Knowledge Injection" strategy. We explicitly engineer physics-based features to decouple the deterministic geometric component of solar irradiance from the stochastic atmospheric component.

- Physical Reference and Clear-Sky Modeling:
  To provide the model with a noise-free reference baseline, a local clear-sky calculation engine was developed (Fig. 2). The calculation sequence begins with determining the Earth-Sun geometric relationship based on the timestamp ($t$). First, the Day of the Year ($n$) is extracted to locate Earth in its orbit (Eq 1). Next, the Solar Declination ($\delta$), representing the tilt of the Earth's axis, is calculated based on the day of the year (Eq 2). These parameters are combined with the local latitude ($\phi$) and hour angle ($\omega$) to derive the Solar Zenith Angle ($\theta_z$), which determines the optical path length (Eq 3).

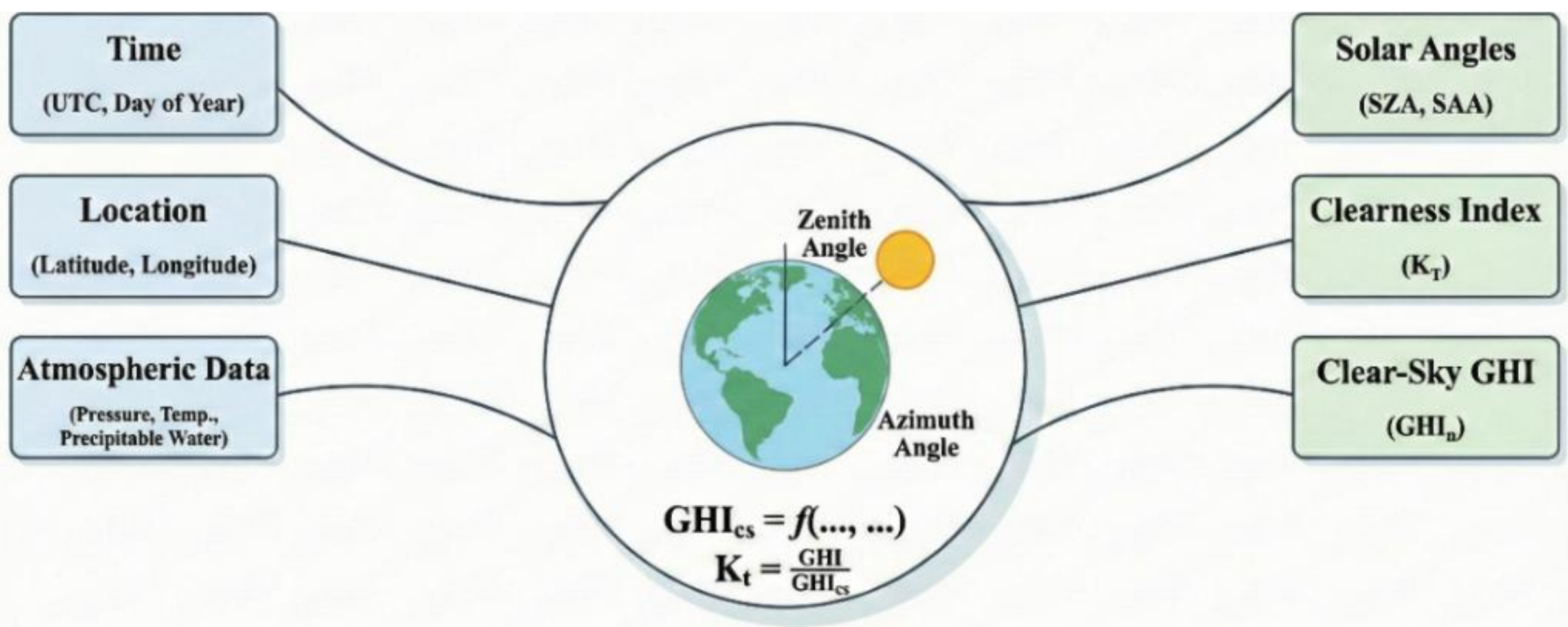

**Fig. 2.** Conceptual diagram of the Physics Calculation Engine, illustrating the derivation of solar angles and clear-sky irradiance from raw inputs.

$$d_n = \text{Day Of Year } (t) \tag{1}$$

Next, the Solar Declination ($\delta$), representing the tilt of the Earth's axis, is calculated as:

$$\delta = 23.45 \cdot \sin\left(360 \cdot \frac{d_n + 284}{365}\right) \tag{2}$$

These parameters are combined with the local latitude ($\phi$) and hour angle ($\omega$) to derive the Solar Zenith Angle ($\theta_z$) which determines the optical path length:

$$cos(\theta_z) = sin(\phi)sin(\delta) + cos(\phi)cos(\delta)cos(\omega) \tag{3}$$

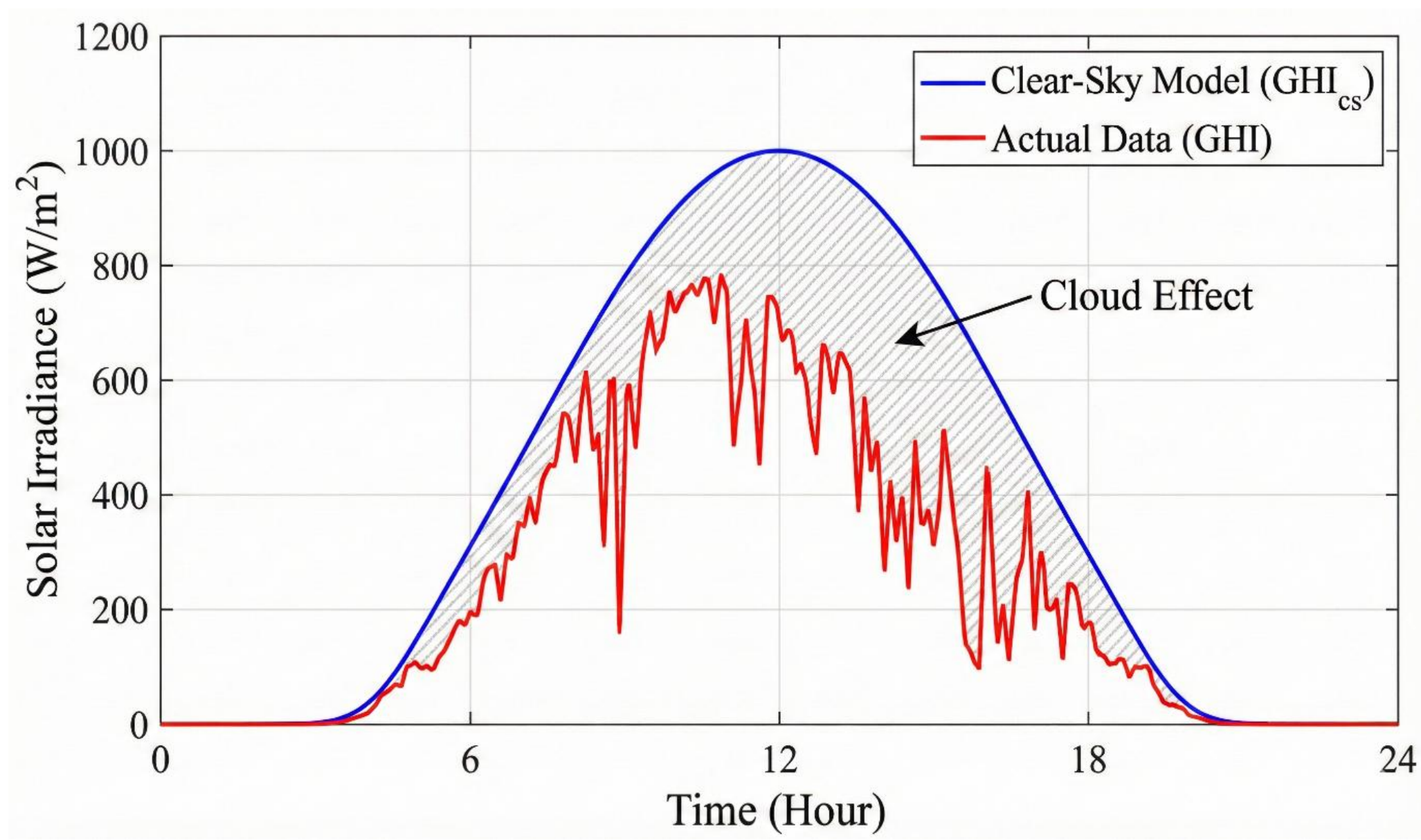

**Fig. 3.** Comparison between the theoretical Clear-Sky model $\left(GHI_{\{cs\}}\right)$ and actual measured GHI, visually defining the stochastic "Cloud Effect" as the deviation from the geometric baseline.

- To estimate the theoretical Clear-Sky GHI ($GHI_{cs}$), we adopted the Kasten-Czeplak formulation. The baseline irradiance is calculated by combining the solar constant $I_{sc}$, an Earth-Sun distance correction factor that accounts for the orbital eccentricity using the day of the year $d_n$, the clear-sky transmission coefficient tuned to 0.7, and an altitude correction factor. Furthermore, a mathematical function is utilized to ensure non-negative values during nighttime (Eq 4). The deviation of the actual measured Global Horizontal Irradiance from the theoretical Clear-Sky model visually defines the stochastic cloud effect. This direct comparison effectively isolates the unpredictable atmospheric variations from the deterministic geometric baseline (Fig. 3).

$$GHI_{cs} = I_{sc} \cdot \left(1 + 0.033\left(\frac{cos360\, d_n}{365}\right)\right) \cdot \max(0, cos\theta_z) \cdot \tau \cdot f_{alt} \quad (4)$$

- Derived Atmospheric Indicators: Based on the physical reference, two critical indicators were derived. The Clearness Index ($K_T$) acts as a cloud filter, isolating the atmospheric transmittance from the solar geometry. It is calculated as the ratio of measured to theoretical irradiance. To ensure numerical stability during nighttime when values approach zero, a constant value was added to the denominator as per the implementation (Eq 5).

$$K_T = \frac{GHI_{meas}}{GHI_{cs} + 1} \quad (5)$$

- Additionally, to capture the dynamic stability of the weather, a Volatility Index ($V_t$) was computed as the rolling standard deviation of the clearness index over a preceding window. This measure alerts the model to rapidly changing conditions, such as fast-moving scattered clouds (Eq 6).

$$\mathrm{V_t} = \sigma(\mathrm{K_T})_{\mathrm{t-w:t}} \quad (6)$$

### 2.4. Mathematical Transformations and Temporal Embedding

To ensure numerical stability and maximize the learning efficiency of the deep neural network, the raw input features underwent specific mathematical transformations before being fed into the model.

1. Logarithmic Transformation for Skewed Distributions:

Meteorological variables, particularly solar irradiance components, typically exhibit right-skewed distributions with heavy tails. Feeding such data directly can lead to gradient instability during backpropagation. To mitigate this, a logarithmic transformation was applied to the wide-range irradiance variables (GHI, DNI, DIFF, and ClearSky). Additionally, to standardize the input space and reduce the impact of outliers, the same transformation was applied to Ambient Temperature ($T_{\{amb\}}$), Satellite Clearness Index ($KT_{\{sat\}}$), and Wet Bulb Temperature ($T_{\{wet\}}$).

$X_{log} = \ln(1 + X_{raw})$ bias term $(+1)$ ensures the function is defined for zero-values (e.g., nighttime irradiance) while maintaining the monotonicity of the data [1]. This transformation approximates a Gaussian distribution, reducing the impact of outliers and allowing the Adam optimizer to converge faster by normalizing the weight updates.

2. Cyclical Time Encoding:

Cyclical Time Encoding: Standard ordinal encoding of time introduces a numerical discontinuity where the last hour of the day and the first hour of the next are perceived by the network as far apart despite being temporally adjacent. To preserve the continuous cyclic nature of diurnal and seasonal patterns, we projected time into a two-dimensional coordinate system using sinusoidal transformations, as illustrated in Fig. 4.

**Fig. 4.** Cyclical Time Encoding (Daily and Annual) for Preserving Temporal Continuity.

The first set of transformations captures the diurnal or hourly cycle. Both sine and cosine components are calculated by multiplying two pi by the ratio of the current hour $h$ to the total hours in a day, which is 24 (Eq 7).

The second set of transformations captures the seasonal or daily cycle. Similarly, the sine and cosine components are computed by multiplying two pi by the ratio of the current day $d$ to the total days in a solar year, represented as 365.25 to accurately account for leap years (Eq 8). Combining both temporal representations guarantees that the distance between the end of one cycle and the beginning of the next is correctly interpreted as a smooth geometric change by the neural network.

- $Diurnal\ Cycle\ (Hourly)$:

$$H_s = sin\left(2\pi \cdot \frac{h}{24}\right), H_c = \cos\left(2\pi \cdot \frac{h}{24}\right) \quad (7)$$

- $easonal\ Cycle\ (Daily)$:

$$D_s = sin\left(2\pi \cdot \frac{d}{365.25}\right),\ D_c = cos\left(2\pi \cdot \frac{d}{365.25}\right) \quad (8)$$

This embedding ensures that the transition from 23:00 to 00:00 (or Dec 31 to Jan 1) involves a smooth geometric change, preventing logic breaks in the time-series analysis [3].

3. Final Input Vector Composition:

The final input vector $X_t$ at each time step consists of 15 features, categorized in Table 2.

**Table 2.** Composition of the hybrid input vector

| Description | Features (Units/Scale) | Category |
|---|---|---|
| The primary variable to be forecasted (Solar Irradiance). | $\text{GHI}\left(\frac{\text{W}}{\text{m}^2}\right)$ | Historical & Target |
| Captures moisture content, saturation levels, and ambient thermal state (Temp is Log-scaled). | Humidity (RH), Dew Point $\left(T_{\{dew\}}\right)$, Temp $\left(T_{\{amb\}}\right)$ | Atmospheric State |
| Raw solar radiation components representing beam and scattered irradiance. | Direct (DNI), Diffuse (DHI) | Radiative Components |
| Replaces wind speed; indicates evaporative cooling potential and thermal comfort. | Wet Bulb Temperature $\left(T_{\{wet\}}\right)$ | Thermodynamic State |
| Theoretical baselines and dual cloudiness indices (Calculated & Satellite-derived.( | ClearSky, $KT_{\{calc\}}$, $KT_{\{sat\}}$ (Log), Volatility | Physics-Informed |
| Cyclic encoding for Hour and Day to capture seasonality and diurnal cycles. | Sin/Cos(H, D) | Temporal Embedding |

**2.5 Data Normalization and Tensor Construction**

Before feeding the data into the deep learning model, the engineered features undergo structural and statistical transformations to ensure training stability and compatibility with the CNN-BiLSTM architecture.

1. Leakage-Proof Normalization Strategy:

Deep neural networks are highly sensitive to the scale of input data. To prevent gradient explosion and ensure faster convergence, all input features were standardized using Z-score normalization. This mathematical technique scales the input data by subtracting the mean from each data point and dividing the result by the standard deviation, effectively transforming the data to have a mean of zero and a standard deviation of one (Eq 9).

$$z = \frac{x - \mu_{train}}{\sigma_{train}} \tag{9}$$

Crucially, to rigorously prevent data leakage, the statistical parameters, specifically the mean $\mu$ and standard deviation $\sigma$, were computed exclusively on the training subset. These frozen parameters were then applied to normalize the validation and testing sets. This strict separation ensures that the model receives no look-ahead information about the future distribution of data in the testing phase, thereby guaranteeing a realistic evaluation of its generalization capability.

2. Sliding Window Sequence Generation:

To capture the temporal dependencies inherent in solar irradiance, the 2D multivariate time series was transformed into 3D tensors using a Sliding Window technique (Fig.5).

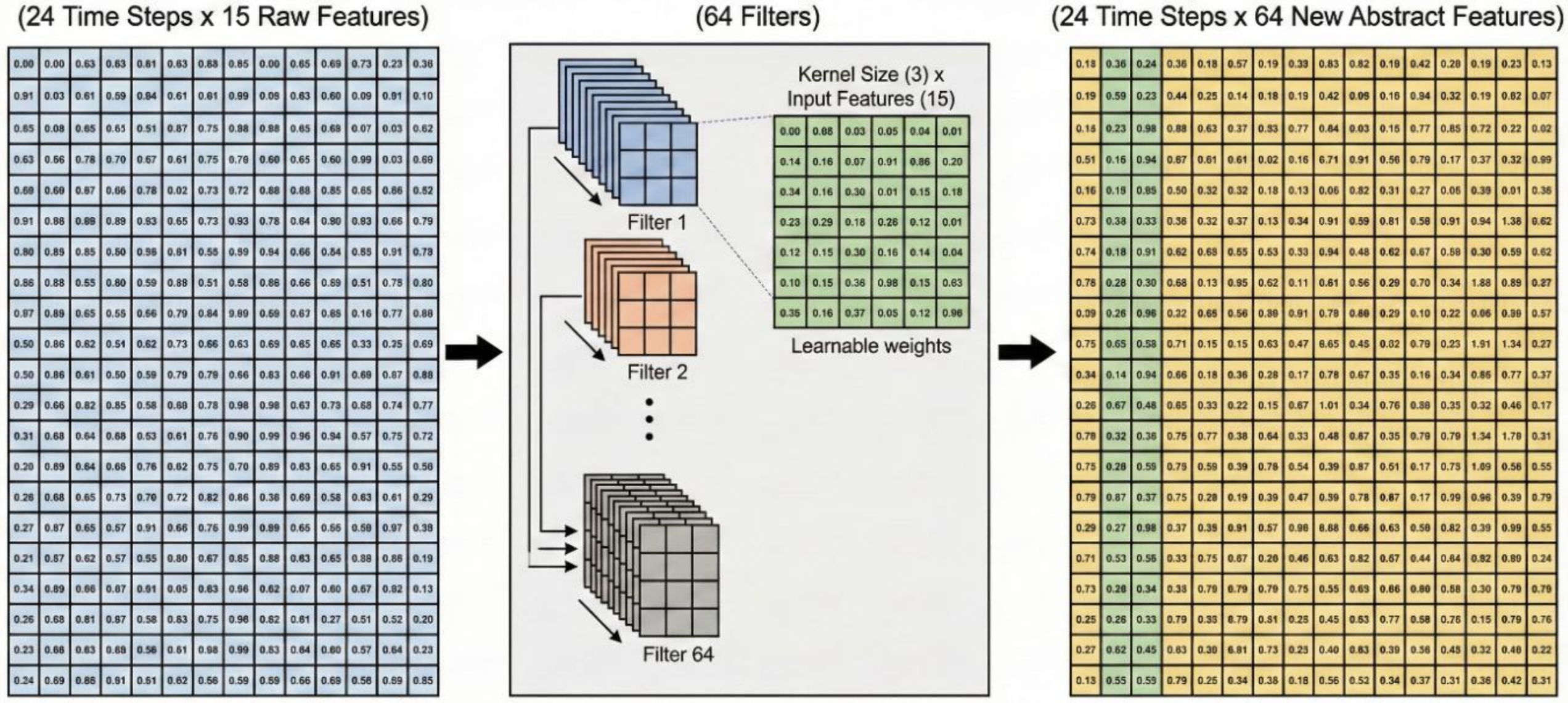

**Fig. 5.** Detailed visualization of the 1D-CNN feature extraction process, showing the 64 filters and the resulting abstract feature maps from the input vector.

The forecasting problem is formulated as a supervised Sequence-to-One regression task. We defined a look-back window size of w = 24 hours to encompass a full diurnal cycle, allowing the model to analyze the complete context of the previous day before making a prediction.

Mathematically, for each time step $t$, the input tensor $X_t$ and target $Y_t$ are constructed as:

- Input Sequence: $X_t = \{x_{t-23}, x_{t-22}, \ldots, x_t\} \in R^{24\times15}$
- Target: $Y_t = GHI_{t+1}$

The window slides forward with a step size of 1 hour, generating samples that map the historical context of the past 24 hours to the GHI value of the immediate next hour $(t+1)$. This structure is ideal for operational, real-time control applications where decisions (like pump scheduling) depend on the immediate future state.

### 2.6 Proposed Hybrid CNN-BiLSTM Architecture

The proposed forecasting system employs a deep hybrid architecture designed to capture both the high-frequency local variations (e.g., passing clouds) and the long-term temporal dependencies (e.g., seasonal trends) of solar irradiance. As illustrated in Fig. 6, the network is composed of three sequential processing blocks:

1. Spatial Feature Extraction (CNN Block):

The input tensor first passes through a 1D-Convolutional Neural Network (1D-CNN) layer. Unlike standard dense layers, the CNN utilizes 64 learnable filters (kernels) that slide across the 24-hour time window. This operation allows the model to extract "local patterns" and rapid weather fluctuations invariant to their position in time.

- Batch Normalization: Applied immediately after convolution to mitigate internal covariate shift, ensuring stable gradient propagation.
- ReLU Activation**:** Introduces non-linearity to capture complex relationships.

2. Temporal Context Learning (BiLSTM Block):

The extracted feature maps are then fed into the core recurrent component: a Bidirectional Long Short-Term Memory (BiLSTM) layer configured with 210 hidden units.

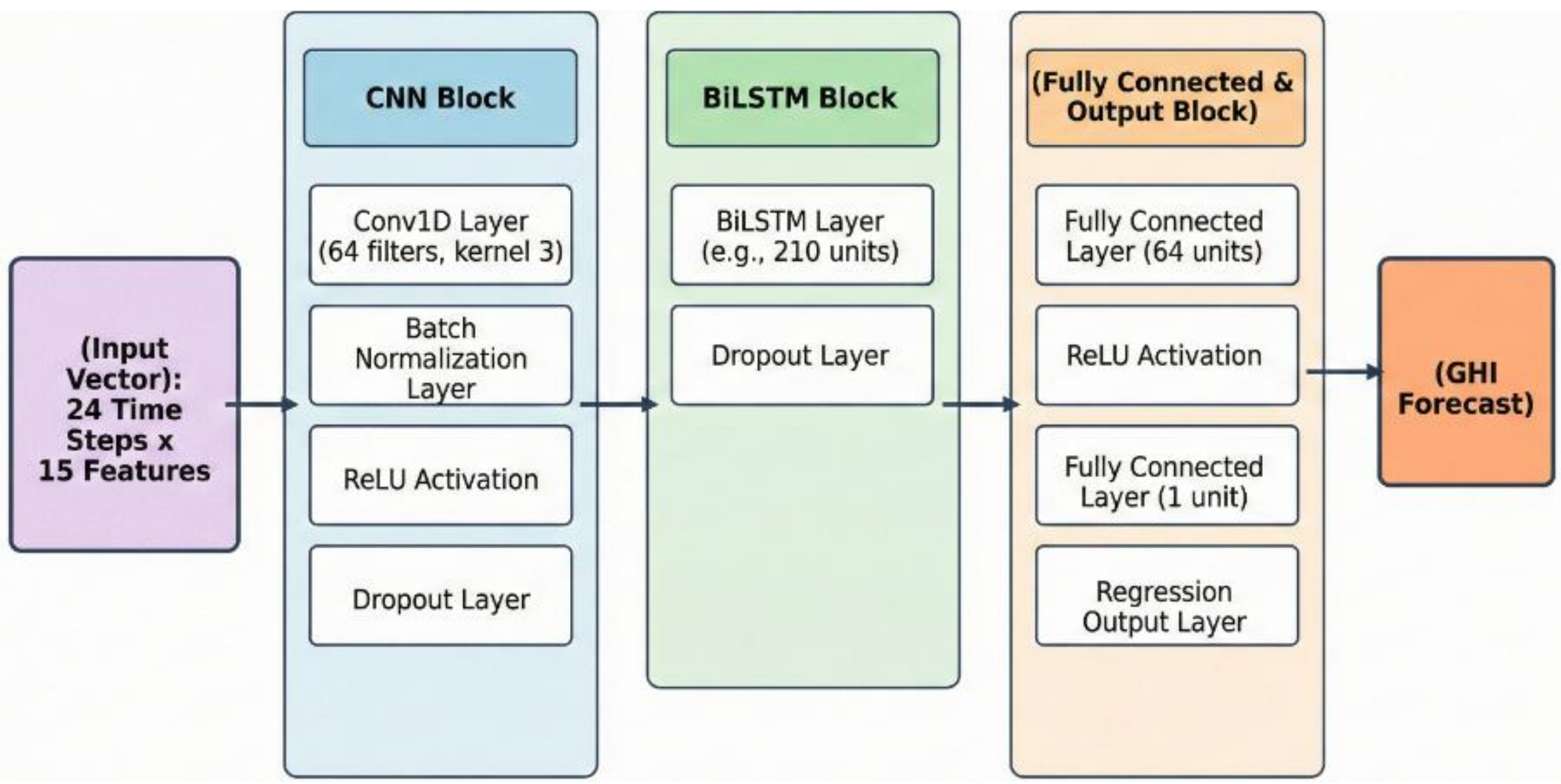

**Fig. 6.** High-level block diagram of the proposed hybrid architecture, showing the sequential flow from the Input Layer through CNN and BiLSTM blocks to the final forecast

Unlike standard LSTMs that process data in one direction, the BiLSTM processes the sequence in both forward ($t \rightarrow t+n$) and backward ($t+n \rightarrow t$) directions. This mechanism enables the model to access the complete context of the sliding window, effectively linking historical radiation states with future trends.

3. Regression and Output (Dense Block):

The temporal features are flattened and passed through Fully Connected (Dense) layers to synthesize the final prediction. The architecture terminates with a single-unit linear layer that outputs the predicted GHI value ($W/m^2$ ).

4. Computational Complexity:
The complete architecture comprises 11 processing layers with a total of 492,200 trainable parameters. This specific configuration represents an engineered balance between "Model Depth" (required to model stochastic weather complexity) and "Computational Lightness," avoiding the extreme computational cost of Transformer-based models which often exceed 1 million parameters Detailed Layer Specifications

The exact configuration of the proposed Hybrid CNN-BiLSTM model is detailed in Table 3. The architecture was engineered with specific design choices to maximize temporal fidelity:

- Preservation of Temporal Resolution:

The initial 1D-CNN layer utilizes a kernel size of 3 with "same" padding1. This is a critical design choice that maintains the original time-step dimension ($T = 24$) throughout the feature extraction phase, ensuring that boundary information (the first and last hours of the window) is processed with equal weight and preventing the sequence shrinkage typical of valid-padding convolutions.

- Bayesian-Controlled Regularization:

Two Dropout layers are strategically placed after the convolutional block and the recurrent block. Unlike static regularization, the dropout rate is treated as a hyperparameter and dynamically tuned via Bayesian Optimization within the range $[0.1, 0.5]$. This adaptive regularization prevents overfitting by ensuring the network learns robust, distributed representations rather than relying on specific neurons.

**Table 3**. Detailed configuration of the proposed hybrid CNN-BiLSTM architecture

| | **Layer Name** | **Configuration / Parameters** | **Function / Role in Project** | **Output Dimensions (N,D,T)** |
|---|---|---|---|---|
| 1 | Sequence Input | Steps: 24, Features: 15 | Receives the physics-guided multivariate time-series data. | $(N, 15,24)$ |
| 2 | 1D-Convolution | Filters: 64, Kernel Size: 3, Padding: 'same'$1 \times 10^{-5}$ | Extracts local short-term weather patterns and rapid fluctuations. | $(N, 64,24)$ |
| 3 | Batch Normalization | Epsilon: | Stabilizes the learning process and normalizes the CNN output distribution. | $(N, 64,24)$ |
| 4 | ReLU Activation | $f(x) = max(0, x)$ | Introduces non-linearity by suppressing negative values in feature maps. | $(N, 64,24)$ |
| 5 | Dropout (1) | Rate: $\alpha$ (Optimized via BayesOpt) | Prevents overfitting by randomly dropping connections after feature extraction. | $(N, 64,24)$ |
| 6 | BiLSTM | Units: $\beta$ (Optimized), Mode: 'Last' | Captures long-term temporal dependencies and historical context (past & future). | $(N, 2 \times \beta)$ |
| 7 | Dropout (2) | Rate: $\alpha$ (Optimized via BayesOpt) | Enhances model robustness after the recurrent memory layer. | $(N, 2 \times \beta)$ |
| 8 | Fully Connected | Neurons: 64 | Fuses the extracted temporal features into high-level representations. | $(N, 64)$ |
| 9 | ReLU Activation | $f(x) = max(0, x)$ | Activates the dense layer features. | $(N, 64)$ |
| 10 | Output Dense | Neurons: 1 | Produces the final continuous prediction value (GHI). | $(N, 1)$ |
| 11 | Regression Output | Loss Function: RMSE | Computes the error and guides the backpropagation process during training. | Scalar |

### 2.7 Bayesian Hyperparameter Optimization Framework

Deep learning models are notoriously sensitive to the initialization of hyperparameters. Manual tuning (trial-and-error) is often computationally prohibitive and suboptimal. To address this, we employed a Bayesian Optimization strategy to automate the search for the global optimum within a high-dimensional search space.

1. Optimization Logic:

As depicted in the flowchart (Fig.9), the algorithm iteratively updates a probabilistic surrogate model to map the relationship between hyperparameters and the objective function. The search process was guided by the Expected Improvement Plus (EI+) acquisition function, which balances "exploration" (searching in uncertain regions) and "exploitation" (refining promising regions).

2. Search Space and Protocol:

We defined a logarithmic search space for the learning rate and regularization terms to cover multiple orders of magnitude efficiently. The optimization budget was set to 30 independent iterations, with the objective of minimizing the Root Mean Square Error (RMSE) on the validation set. Table 4 details the boundaries of the search space.

3. Optimization Outcome:

The optimization process converged to a configuration that favors high model capacity (210 hidden units) with relatively low regularization (Dropout 10%). This outcome suggests that the physics-guided features provide a strong, clean signal, reducing the need for aggressive regularization to prevent overfitting.

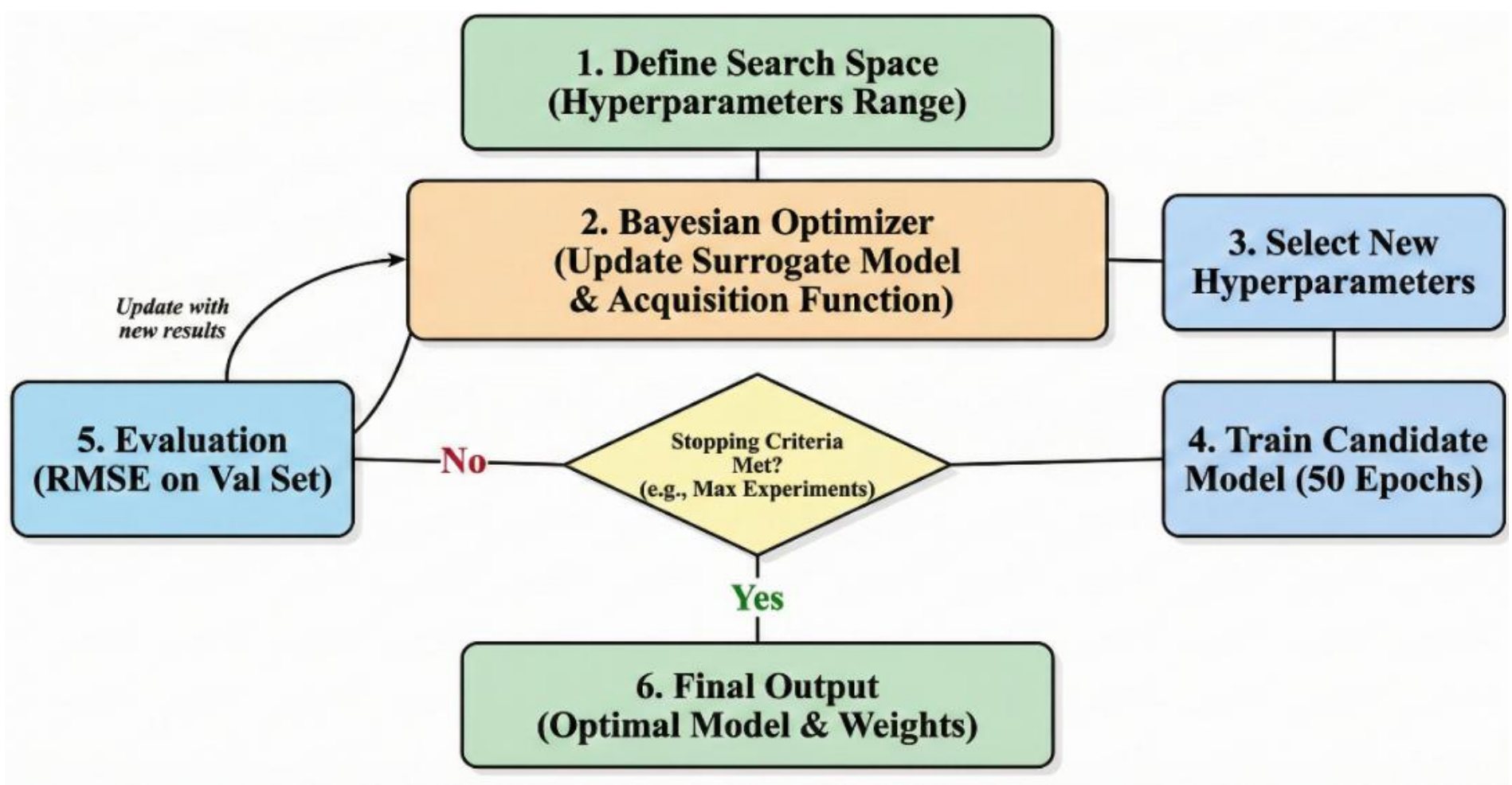

**Fig. 9.** Flowchart of the Bayesian Optimization process using the Expected Improvement Plus (EI+) acquisition function.

**Table 4**. Hyperparameter search space and optimal values

| Hyperparameter | Search Range | Scale | Physical/Structural Role | Optimal Value |
|---|---|---|---|---|
| Initial Learn Rate | $[10^{-4}, 10^{-2}]$ | Log | Controls weight update speed | 0.00175 |
| L2 Regularization | $[10^{-5}, 10^{-3}]$ | Log | Weight Decay (Prevents large weights) | $1.2 \times 10^{-4}$ |
| BiLSTM Units | 100 ,250 | Integer | Temporal memory capacity | 210 |
| Dropout Rate | [0.1, 0.5] | Linear | Prevents neuron co-adaptation | 0.104 |

## 2.8 Training Protocol and Stability Measures

To ensure both computational efficiency during the search phase and convergence stability during the final deployment, a rigorous two-stage training strategy was adopted.

1. Two-Stage Training Strategy:

- Exploration Phase: During the Bayesian Optimization search, candidate models were trained for a restricted budget of 50 epochs. This constraint allowed for a rapid exploration of the hyperparameter space without incurring excessive computational costs.
- Refinement Phase: Once the optimal hyperparameters were locked, the final model was retrained from scratch with an extended horizon of 100 epochs. This ensures the model weights have sufficient time to settle into a deep global minimum.

2. Stability and Regularization Techniques:

As detailed in Table 5, the training process incorporated specific mechanisms to mitigate overfitting and ensure numerical stability:

- Adaptive Optimization: The Adam optimizer was selected for its efficiency in handling sparse gradients and non-stationary objectives, utilizing a mini-batch size of 128.
- Inherent Stability and Gradient Management: Interestingly, despite the high volatility of solar irradiance in tropical regions, the model exhibited inherent numerical stability. Experimental observations revealed that the proposed physics-guided architecture did not require additional stabilization techniques, such as Gradient Clipping. This is attributed to the engineered physical features (e.g., Clear-Sky index), which provided a well-conditioned input space that prevented gradient instability.

- L2 Regularization (Weight Decay): To constrain model complexity, an L2 penalty term was added to the loss function. The regularization coefficient ($\lambda$) was dynamically tuned via Bayesian Optimization. This mechanism actively suppresses large weights, forcing the network to learn smoother, more generalizable patterns rather than memorizing noise.
- Early Stopping: To prevent overfitting, a "patience" mechanism monitored the validation loss. Training was automatically halted if no improvement was observed for 20 consecutive epochs, ensuring the model was saved at its peak generalization performance.

**Table 5.** Training hyperparameters and stability settings

| Parameter | Value / Method | Function |
|---|---|---|
| Optimizer | Adam | Adaptive moment estimation for stable convergence |
| Batch Size | 128 | Balances memory usage and gradient estimation accuracy |
| Max Epochs | 100 | Maximum iterations for the final model |
| Early Stopping | Patience = 20 | Prevents overfitting by monitoring validation loss |
| L2 Regularization | Optimized (λ) | Penalizes complexity to improve generalization |

### 2.9 Implementation Environment and Computational Cost

The proposed framework was implemented using MATLAB (R2024b) supported by the Deep Learning Toolbox. All experiments were conducted on a computing workstation equipped with an Intel Core i5 CPU and 16GB RAM.

Given the computational intensity of the convolutional operations and the sequential BiLSTM calculations, the training process incurred a tangible computational cost. The Bayesian Optimization phase required approximately 5 hours of continuous processing to thoroughly explore the hyperparameter space. While computationally demanding, this investment is justified by the significant gain in model robustness and accuracy compared to lower-cost, manually tuned statistical models.

### 2.10 Post-Processing: Physical Consistency Filter

Although the model achieves high accuracy, regression-based neural networks may occasionally produce slightly negative irradiance values due to mathematical approximations, which is physically impossible since the Global Horizontal Irradiance must always be greater than or equal to zero.

To ensure physical validity, a post-processing ReLU-like filter was applied to the final output. This mathematical function calculates the final output $y_{final}$ by taking the maximum value between zero and the model's raw prediction $\widehat{y_{pred}}$, effectively acting as a logic gate that clamps any negative predictions to zero (Eq 10).

$$y_{\mathrm{final}} = \max\left(0, \widehat{y_{\mathrm{pred}}}\right) \tag{10}$$

This crucial step ensures that the final predicted values remain non-negative, guaranteeing that the control signals sent to the photovoltaic PV system are always physically realizable and logically sound for real-world applications.

### 2.11 Comparative Framework: Ablation and Addition Study

To systematically isolate the contribution of physics-informed inputs versus structural complexity, and to validate the proposed hybrid architecture, a rigorous ablation study was designed. The proposed model is benchmarked against two distinct reference models, as summarized in Table 6.The Standard Baseline Model (Data-Driven Reference):

This model is designed to represent the traditional raw data-driven approach. It relies on standard satellite inputs (including the Satellite Clearness Index $\mathrm{K}_{\{t_{sat}\}}$ provided within the raw dataset) but lacks the Analytical Physics Guidance that links these data to celestial and site-specific geometry. Structurally, this model shares the same backbone (CNN-BiLSTM) as the proposed framework but differs by being restricted to 12 variables only; specifically excluding the three locally-derived physical features: Calculated Clear-Sky GHI ($GHI_{\{cs\}}$), Instantaneous Calculated Clearness Index ($K_{\{t_{calc}\}}$), and Volatility Index (sigma). This exclusion aims to isolate the impact of Added Engineering Knowledge versus merely available raw data. Furthermore, fixed hyperparameters (Manual Tuning) were used to simulate common literature settings and highlight the advantage of intelligent optimization.

2.The Attention-Based Hybrid Model (Complexity Test):

The Attention-Based Hybrid Model for the Complexity Test was intentionally designed with increased structural complexity.This variant features stacked CNN layers, a BiLSTM layer, and a self-attention mechanism, but was trained without Bayesian optimization to represent the common brute-force modeling approach. Despite its higher parameter count, it failed to match the stability of the optimized physics-informed hybrid model, confirming that intelligent physics guidance is more effective than unoptimized structural depth.
The primary objective of this configuration is to test the hypothesis of Complexity Saturation. This model rigorously investigates whether adding computational complexity via attention mechanisms yields better forecasting results than physical guidance. It uses the full physics-enhanced input set consisting of 15 features but employs a significantly heavier architecture.

$$Attention(Q,K,V) = softmax\left(\frac{QK^T}{\sqrt{d_k}}\right)V \tag{11}$$

Regarding the mechanism, the Self-Attention layer is specifically placed after the BiLSTM layer to compute dependencies between time steps using the scaled dot-product formula. The attention output is calculated mathematically by applying a softmax function to the dot product of the Query matrix $Q$ and the transposed Key matrix $K^T$, scaled by dividing it by the square root of the key dimension $d_k$, and finally multiplying that normalized result by the Value matrix $V$ (Eq 11). To maintain the correct architectural flow, the matrices $Q$ representing the Query, $K$ representing the Key, and $V$ representing the Value are all linear projections derived directly from the BiLSTM hidden states.

**Table 6.** Summary of models in the ablation and complexity study

| Model ID | Architecture | Input Features | Optimization | Objective |
|---|---|---|---|---|
| Standard Baseline | CNN-BiLSTM | Raw (12 features) | Manual | Measure Physics Impact |
| Attention-Hybrid | CNN -BiLSTM- Attn | Full (15 features) | Manual | Measure Complexity Impact |
| Proposed PI-Hybrid | CNN-BiLSTM | Full (15 features) | Bayesian | Final Optimal Design |

## 2.13 Performance Evaluation Metrics

To quantitatively assess the forecasting accuracy and validate the superiority of the proposed physics-informed framework against the baseline models, three standard statistical metrics were adopted. In all the following mathematical formulations, $N$ represents the total number of data samples, $y_{act}$ denotes the actual measured Global Horizontal Irradiance, $y_{pred}$ is the predicted irradiance value generated by the model, and $\overline{y}_{act}$ stands for the mean of the observed actual values.

The first metric is the Root Mean Square Error, which serves as the primary criterion for our control-oriented application. It is particularly sensitive to large errors or outliers, making it crucial for evaluating the reliability of the model in avoiding drastic prediction failures that could potentially destabilize the PV system. Mathematically, it is calculated as the square root of the average of the squared differences between the actual and predicted values over the entire dataset (Eq 12).

$$\mathrm{RMSE} = \sqrt{\frac{1}{\mathrm{N}}\sum_{\mathrm{i}=1}^{\mathrm{N}}\left(\mathrm{y}_{\mathrm{act}} - \mathrm{y}_{\mathrm{pred}}\right)^2} \tag{12}$$

The second metric is the Mean Absolute Error. This metric is calculated to measure the average magnitude of errors in a set of predictions, thereby providing a linear representation of the general accuracy of the model. It is simply computed as the average of the absolute differences between the actual measurements and the predictions (Eq 13).

$$MAE = \frac{1}{N}\sum_{i=1}^{N}\left|y_{act} - y_{pred}\right| \tag{13}$$

Finally, the Coefficient of Determination $R^2$ is utilized to determine the proportion of variance in the dependent variable that is predictable from the independent variables. This metric effectively indicates the overall goodness of fit of the model and is derived by subtracting the ratio of the sum of the squared prediction errors to the total sum of squares from one (Eq 14).

$$R^2 = 1 - \frac{\Sigma\left(y_{act,i} - y_{pred,i}\right)^2}{\Sigma\left(y_{act,i} - \overline{y_{act}}\right)^2} \tag{14}$$

# 3. Experimental results and discussion

To rigorously evaluate the proposed framework, the model was tested on a designated hold-out dataset. The following analysis focuses on the results obtained for the testing year 2023, which represents the model's peak performance benchmark under standard operating conditions. (A multi-year robustness analysis covering 2020–2024 is presented later in Section 3.8).

## 3.1 Statistical Accuracy and Error Analysis

The predictive fidelity of the model was first assessed through a comprehensive statistical diagnosis of the outputs against the ground truth GHI data.

1. Regression Analysis and Correlation:

Fig .8 illustrates the linear regression plot of predicted vs. measured irradiance. The data points exhibit a tight, dense clustering along the 45-degree identity line, indicating an exceptional agreement between the model's output and reality. The model achieved a Coefficient of Determination $(R^2)$ of 0.9969, implying that the proposed physics-informed architecture successfully explains over 99.69% of the variance in the solar irradiance data. Notably, the linearity is maintained across the entire spectrum, with no observable bias in either low-irradiance (sunrise/sunset) or high-irradiance (noon) regions.

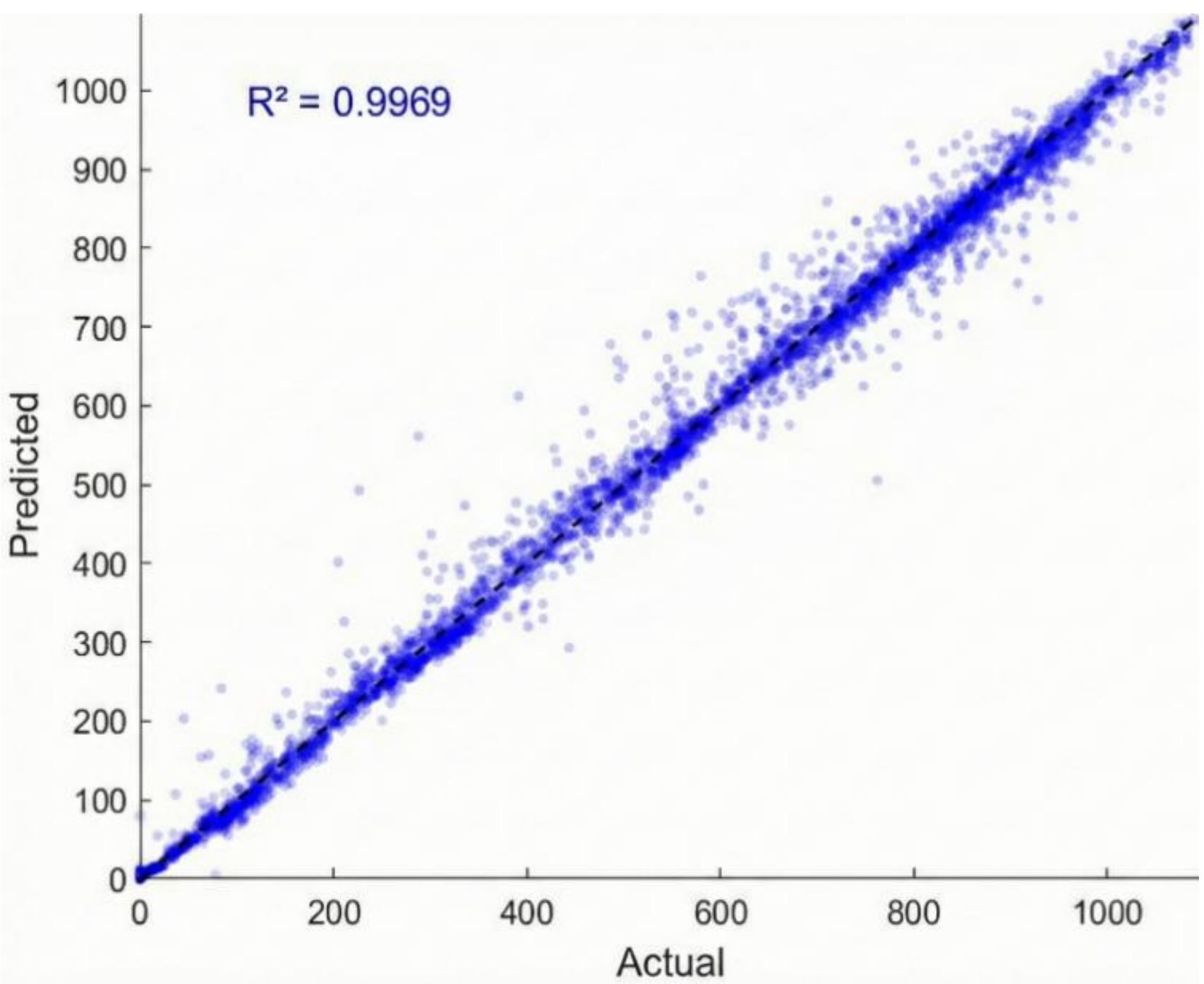

**Fig. 8.** Linear regression analysis between predicted and measured GHI, demonstrating exceptional predictive fidelity with a Coefficient of Determination $(R^2 = 0.9969)$.

2. Error Distribution Characteristics:

To verify the unbiased nature of the predictions, the frequency distribution of forecasting errors is presented in Fig 9. The histogram reveals a near-perfect Gaussian (Normal) Distribution centered precisely at zero with a narrow standard deviation. This symmetry confirms that the model is free from systematic errors $Bias \approx 0$) (and that the minor deviations observed are attributable to the natural stochastic noise inherent in sensor measurements rather than algorithmic deficiencies.

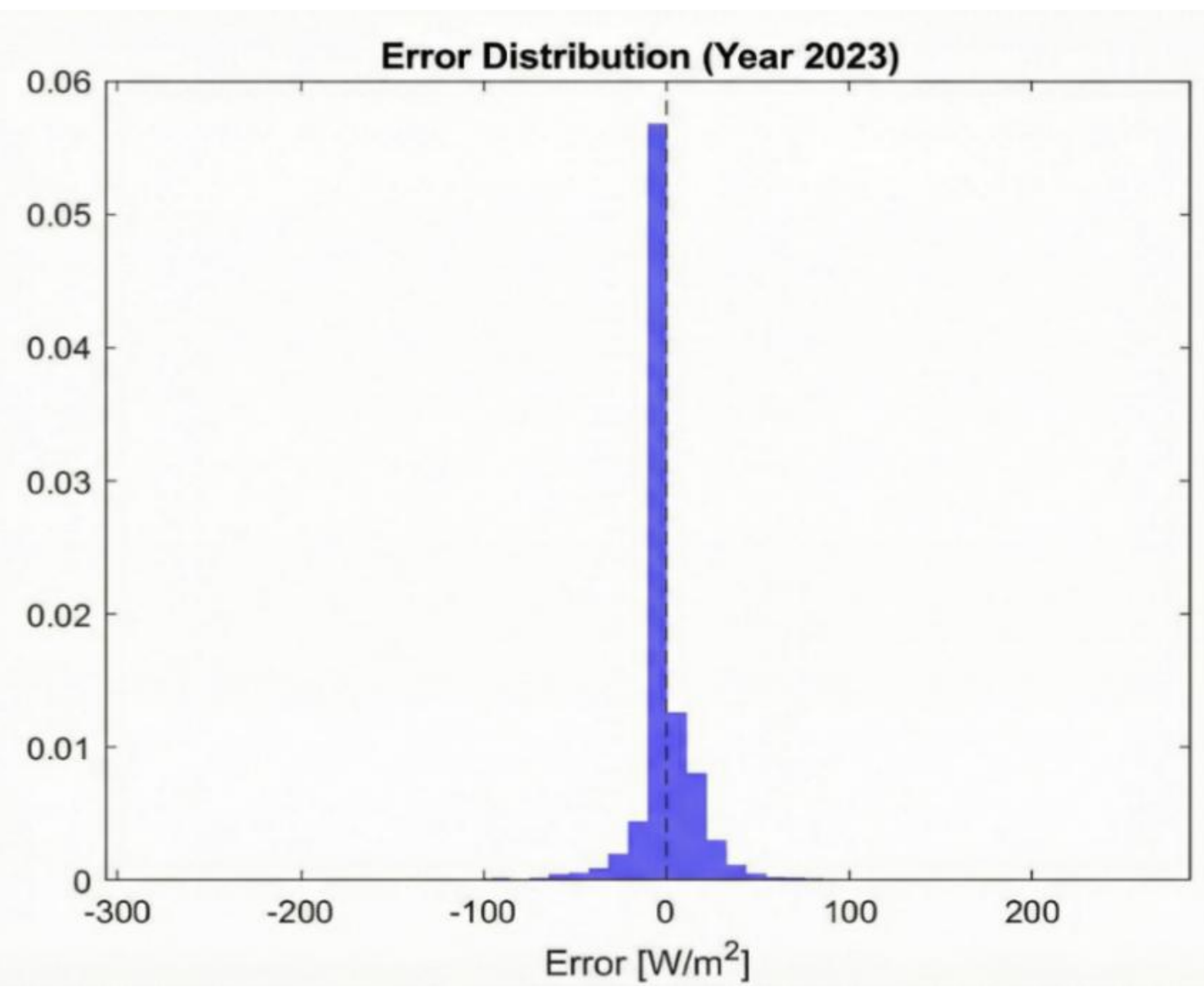

**Fig. 9.** Frequency distribution (Histogram) of forecasting errors exhibiting a near-perfect Gaussian profile centered at zero, indicating an unbiased model.

3. Residual Diagnostics and Homoscedasticity:

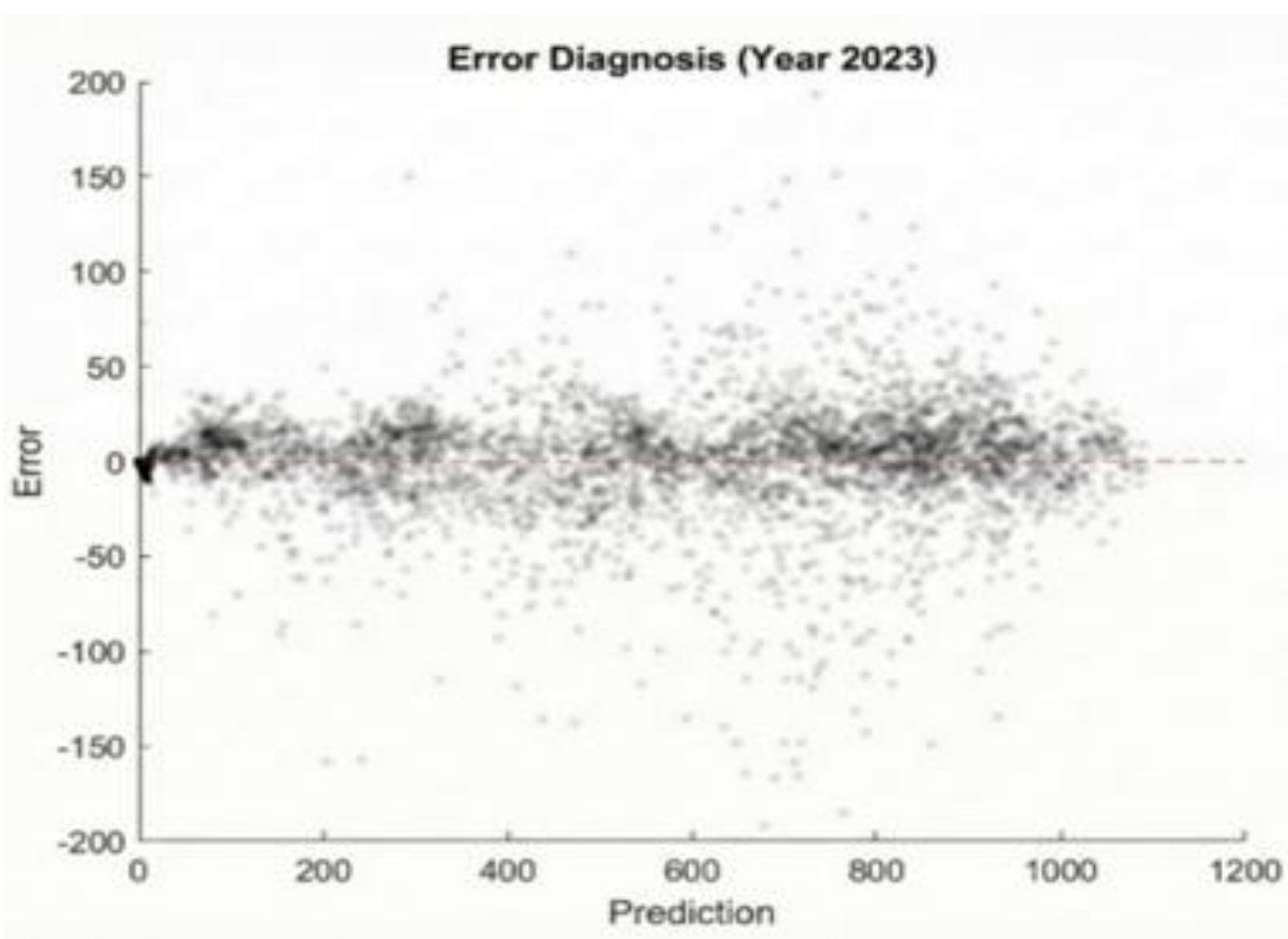

**Fig. 10.** Residual diagnostics showing random error dispersion over time, confirming homoscedasticity and the extraction of all deterministic patterns

Further validation was conducted using the residual plot in Fig 10, which displays the forecast error over time. The residuals show a random, uniform dispersion around the zero line without forming any discernible geometric patterns or trends. This behavior confirms the assumption of Homoscedasticity (constant variance). It indicates that the model has successfully extracted all deterministic physical and temporal patterns from the data, leaving behind only irreducible White Noise, which is the hallmark of a theoretically optimal forecasting model.

## 3.2 Temporal Tracking and Dynamic Response Analysis

For real-time applications such as Model Predictive Control (MPC), statistical accuracy alone is insufficient; the ability to capture instantaneous fluctuations without time lag is paramount.

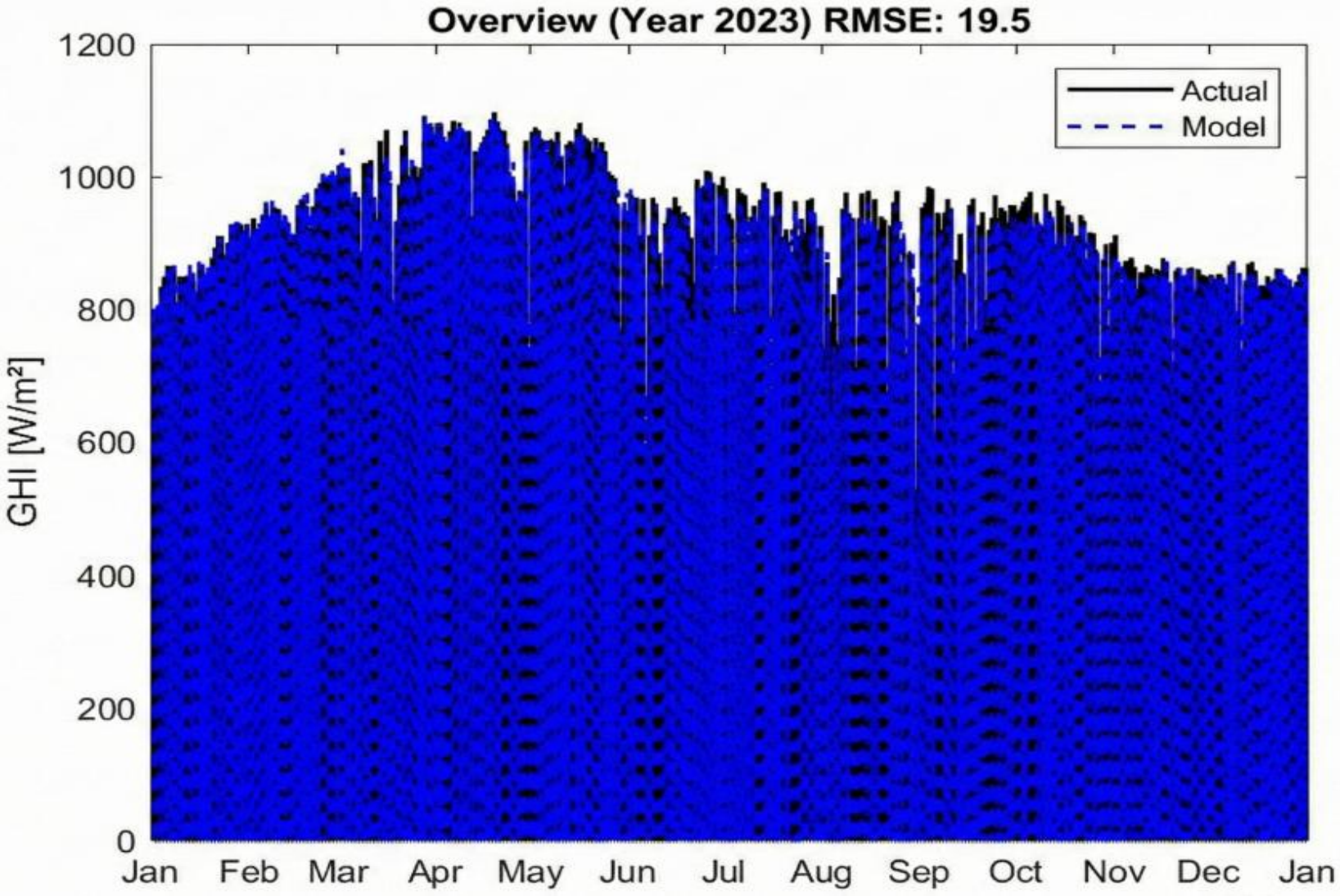

**Fig. 11.** Time-series comparison between predicted and measured GHI over consecutive diurnal cycles, highlighting the model's precise phase synchronization and geometric awareness.

1. Diurnal Cycle Fidelity:

Fig. 11 presents the comprehensive time-series comparison between the predicted and actual GHI over consecutive daily cycles. The model exhibits near-perfect phase synchronization, where the predicted curve (dashed blue) tightly overlays the measured data (solid black). Crucially, the model correctly anticipates the sunrise and sunset moments driven by the geometric inputs (Clear-Sky GHI), eliminating the "shift" often seen in purely autoregressive models.

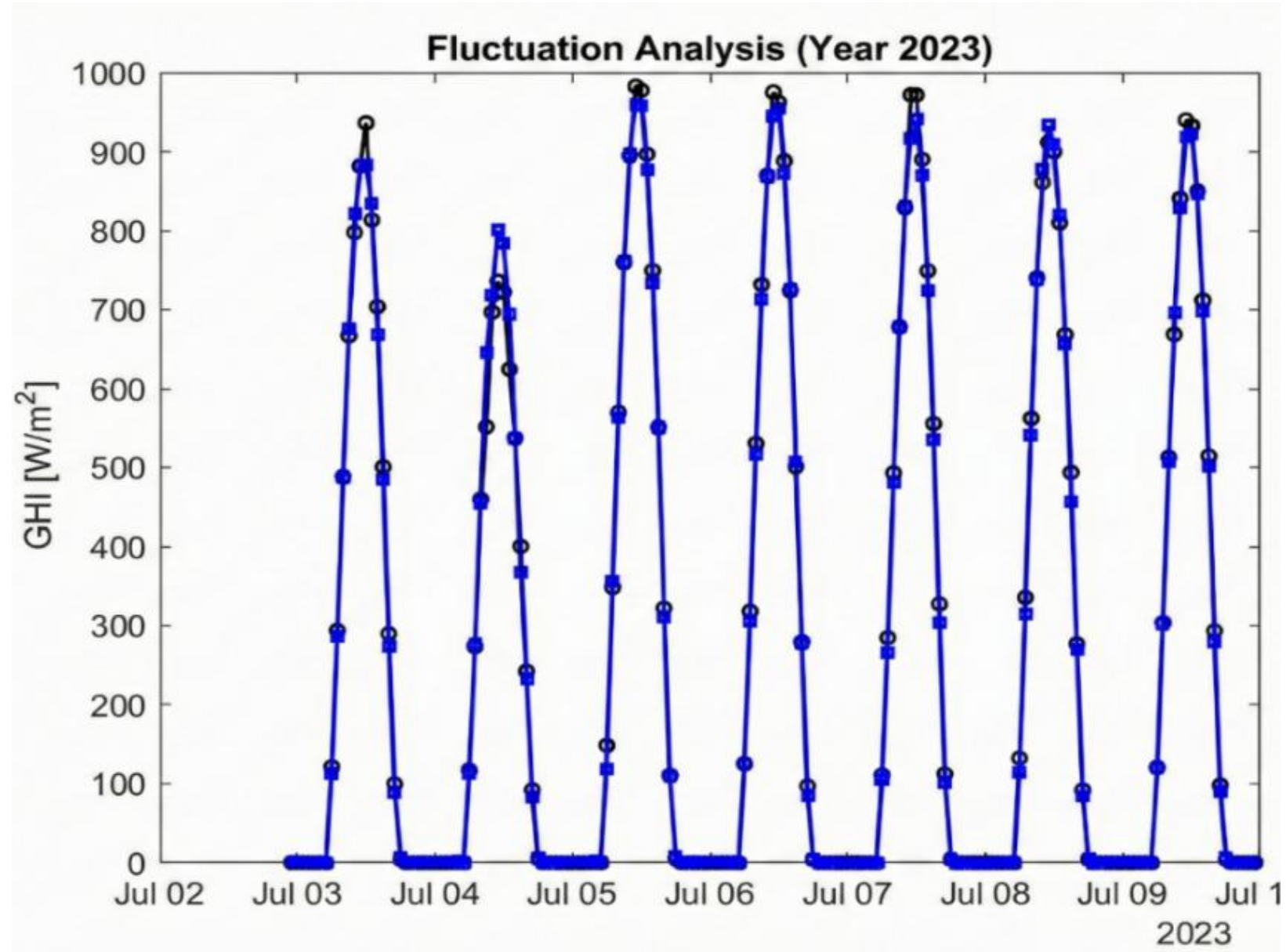

**Fig. 12.** Magnified view of the model's dynamic response during intermittent cloud cover, demonstrating exceptional sensitivity to sharp ramp-down events and rapid cloud-induced transients.

2. Response to High-Volatility Conditions:

To stress-test the model's dynamic capabilities, Fig. 12 provides a magnified view during a period characterized by intermittent cloud cover. The results demonstrate exceptional sensitivity to rapid atmospheric changes. The model successfully tracks sharp ramp-down events (sudden drops in irradiance due to passing clouds) and subsequent recoveries with minimal latency.

3. Physical Interpretation:

This responsiveness—specifically the absence of phase lag during cloud transients—is attributed to the integration of dynamic precursors like Surface Pressure ($P_S$) and Wind Speed ($WSC$) in the feature set. These variables act as early indicators of atmospheric instability, enabling the network to "sense" incoming disturbances rather than merely defaulting to a persistence-based forecast (i.e., repeating the previous hour's value).

### 3.3 Comparative Failure Analysis: Proposed vs. Standard Baseline

To evaluate the added value of the physics-informed features, a direct comparison was conducted against the Standard Baseline (CNN-BiLSTM without physics inputs). This analysis highlights the critical limitations of purely data-driven approaches.

1. Dispersion and Geometric Blindness:

Fig. 13 presents the scatter plot for the Standard Baseline model. In contrast to the tight clustering observed in the proposed model, the baseline exhibits significant dispersion and a much lower $R^2$. This "Geometric Blindness" is particularly evident during clear-sky days, where the baseline fails to follow the deterministic solar arc, resulting in scattered predictions even under stable conditions.

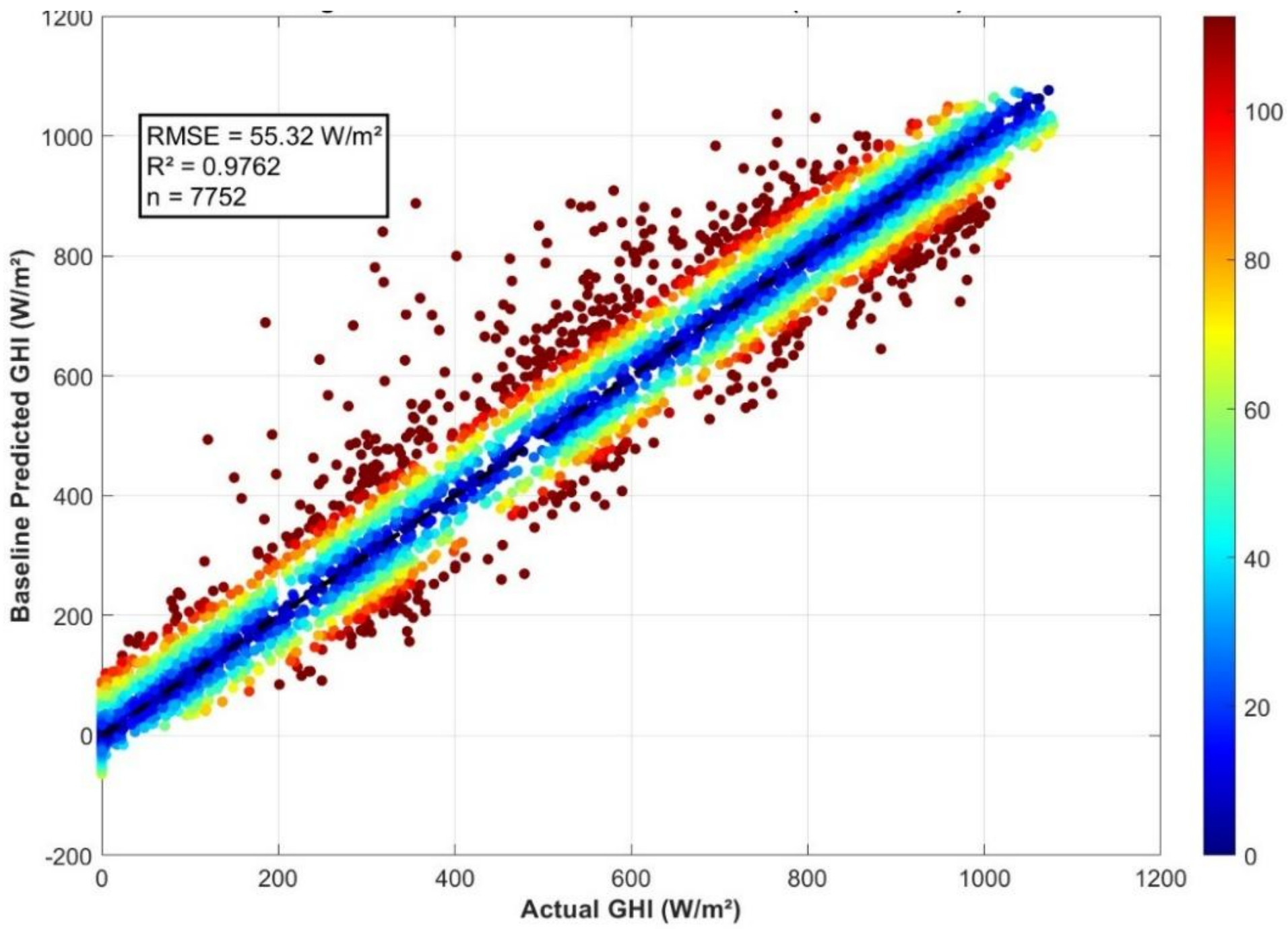

**Fig. 13.** Scatter plot of the Standard Baseline Model (Data-Driven), highlighting high dispersion and the "geometric blindness" caused by the absence of physical precursors.

2. The Phase Lag Phenomenon:

One of the most critical failures identified in the baseline model is the "Phase Lag" or temporal shifting, as visualized in Fig. 14. Because the baseline relies purely on historical lags without understanding solar geometry, it tends to "guess" cloud transients after they have already occurred. This results in a persistent time-shift in the forecast, which is detrimental for real-time grid control.

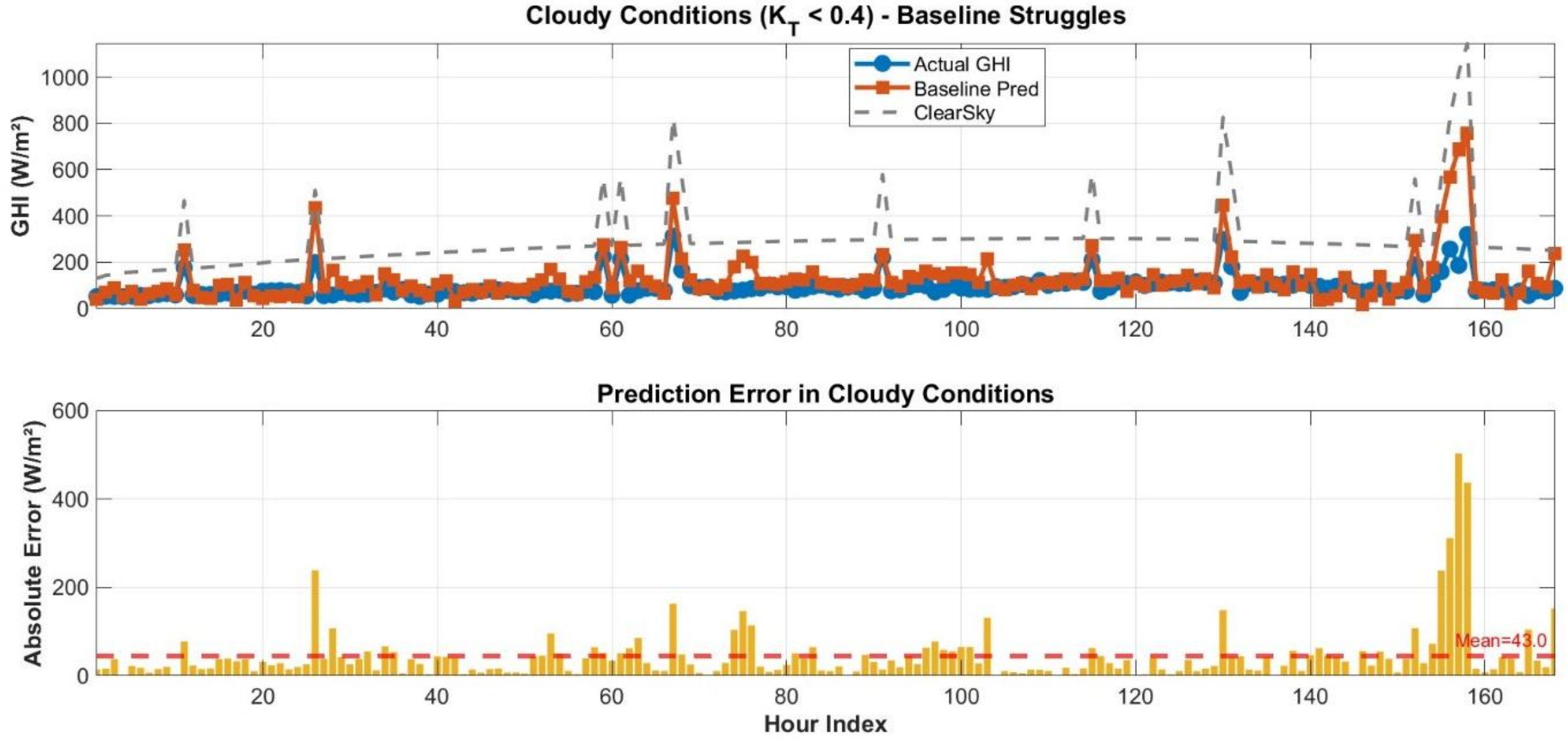

**Fig. 14.** Illustration of the "Phase Lag" phenomenon in the baseline model; the predicted curve exhibits a temporal shift (delay) relative to the ground truth during rapid fluctuations.

3. Instability during Transitional Periods and High Volatility

A. Transition Failures (Sunrise and Sunset): As illustrated in Fig. 15, the standard baseline model exhibits significant forecasting errors during the early morning and late afternoon. This failure is directly linked to the absence of "Solar Geometry" inputs; without a clear-sky reference, the model cannot distinguish between a geometric sunset and a stochastic cloud event, leading to erratic and non-physical predictions.

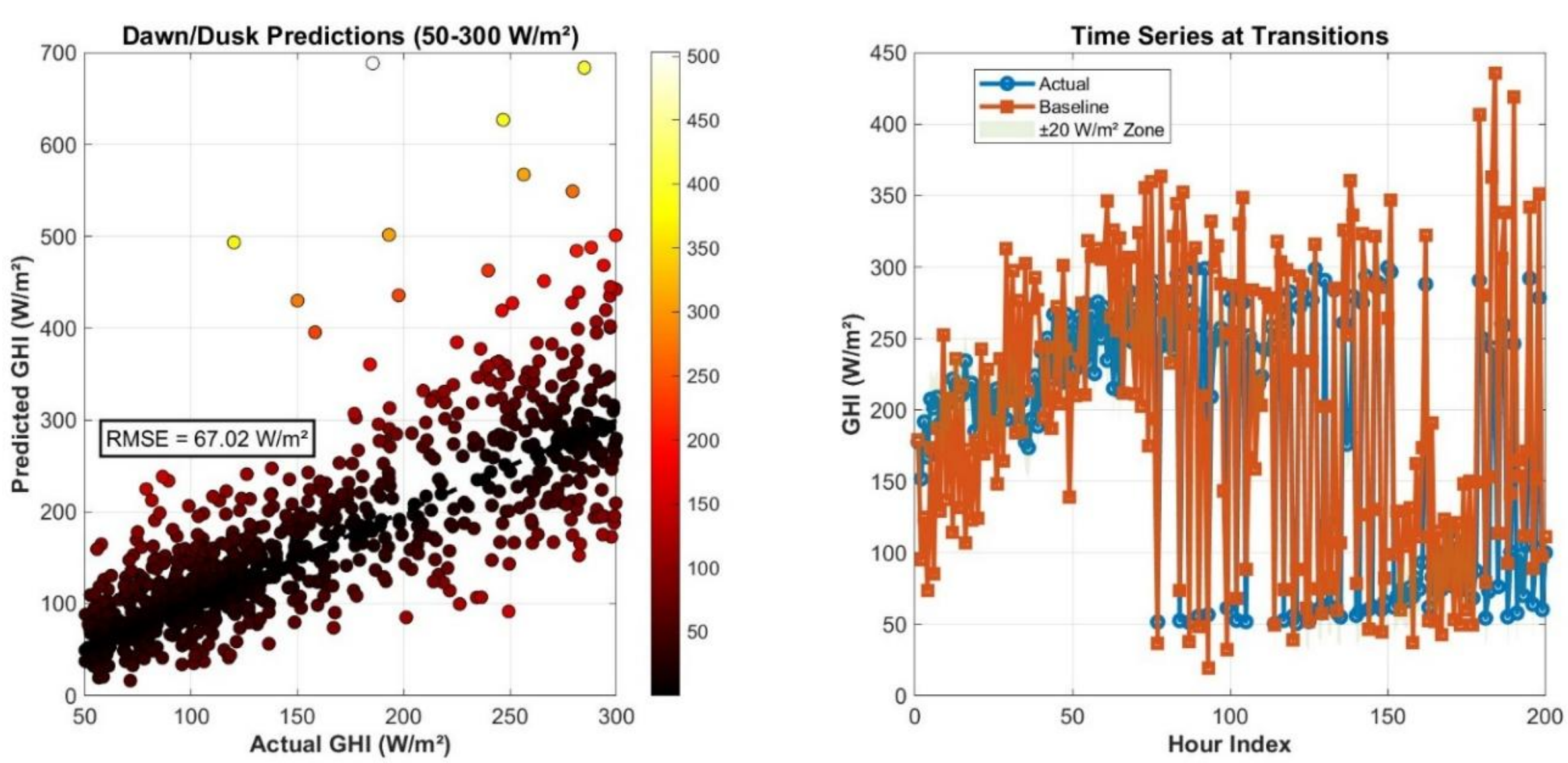

**Fig. 15.** Large forecasting errors of the standard baseline model during transitional periods (sunrise and sunset) due to the lack of solar geometry features.

## 3.4 Volatility Amplitude Mismatch

Furthermore, Fig. 16 highlights the baseline's inability to track high-volatility fluctuations. The standard model tends to exhibit an over-smoothed behavior, effectively "averaging" the irradiance rather than following the dynamic peaks and valleys of the actual signal. This inability to reach the true atmospheric amplitude renders the data-driven baseline unreliable for systems requiring precise ramp-rate control.

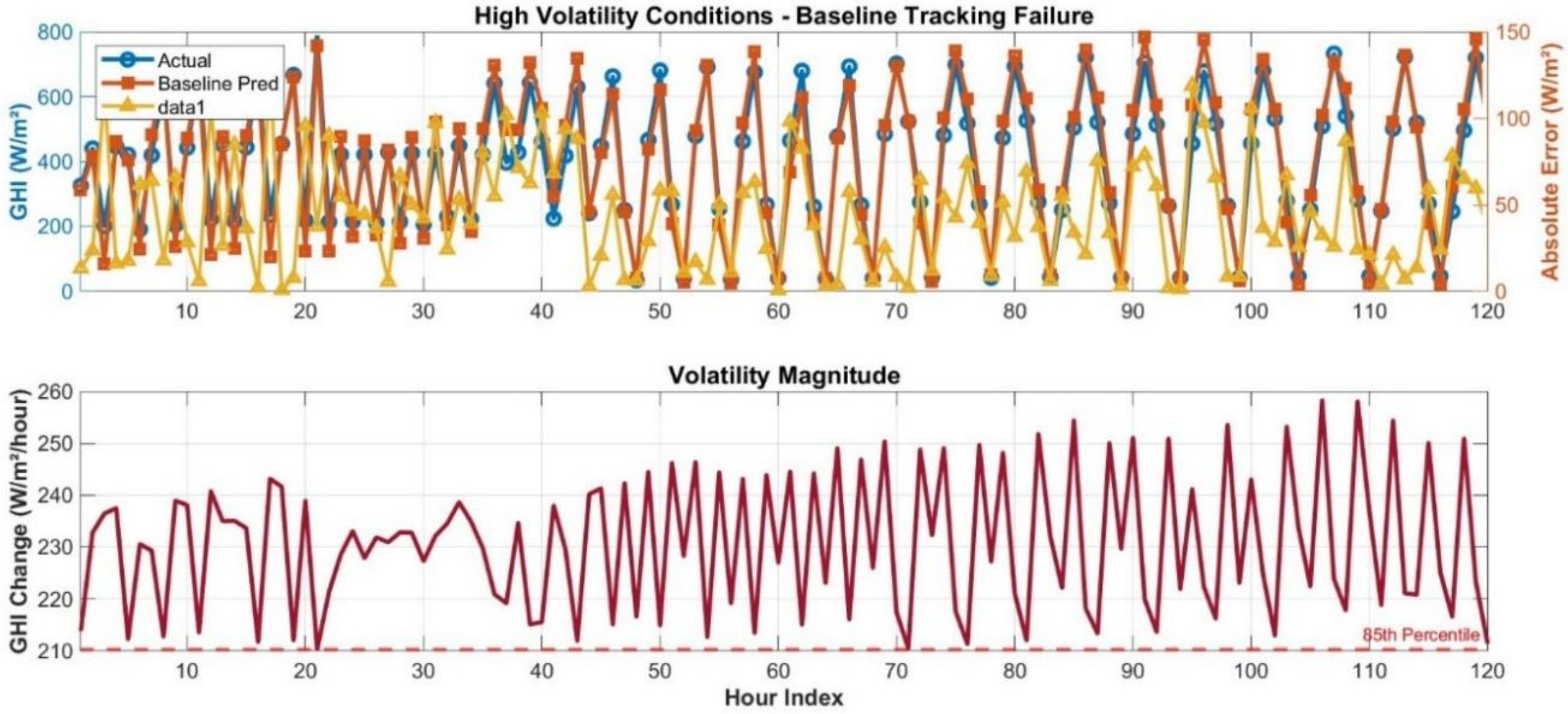

**Fig. 16.** The baseline model's failure to capture high-volatility atmospheric states, demonstrating "over-smoothed" predictions that miss real peaks and valleys.

## 3.5 Benchmarking Against State-of-the-Art

To position the proposed framework within the global research landscape, a comprehensive comparative analysis was conducted against cutting-edge models published in high-impact journals between 2021 and 2025. The comparison focuses on balancing statistical accuracy with architectural efficiency and identifying research gaps in arid-region forecasting.

Recent studies have heavily relied on complex decomposition and generative techniques to improve accuracy. For instance, Li et al. (2023) proposed a CEE-WGAN-LSTM framework for solar irradiance forecasting in semi-arid conditions, while Hou et al. (2023) integrated attention mechanisms with CNN-LSTM to capture temporal dependencies. While these models demonstrate strong performance, they often incur high computational overhead due to their complex architectures. In contrast, our proposed physics-informed hybrid model achieves a superior RMSE of 19.5 W/m² by embedding explicit physical laws, such as the clear-sky index, directly into the learning process, proving that physical guidance can be more effective than purely data-driven complexity.

A key contribution of this work is challenging the trend of using massive deep learning networks for solar forecasting. Transformer-based models, like those proposed by Al-Ali et al. (2023), and dual-stream architectures, such as that of Alharkan et al. (2023), utilize heavy parameterization to handle non-linearities. While Radzi et al. (2025) recently highlighted the importance of metaheuristic optimization for grid-connected systems, many such models remain computationally expensive for real-time deployment. Our framework addresses this by employing a lightweight CNN-BiLSTM architecture, with approximately 492,200 parameters, optimized via Bayesian optimization, similar to the approach suggested by Herrera-Casanova et al. (2024), but with a specific focus on physical inputs. This design reduces computational cost significantly compared to transformer models, validating our hypothesis that a physics-aware, compact model is the optimal solution for resource-constrained environments.

Interestingly, the ablation study detailed in Section 3.7 revealed that adding self-attention mechanisms, a hallmark of modern architectures, actually degraded performance compared to the physics-guided approach. This counterintuitive finding challenges the complexity-first paradigm and demonstrates that explicit physical constraints act as superior, deterministic attention mechanisms when strong domain knowledge is available.

## 3.6 Verification of Physical Consistency

Before deploying the model for control purposes, it is crucial to verify that its predictions are grounded in physical reality rather than mere statistical correlation.

1. Thermodynamic Correlation:

Fig. 17 visualizes the correlation between Ambient Temperature ($T_{\{amb\}}$) and predicted Solar Irradiance ($GHI$). The heatmap reveals a distinct physical pattern: high irradiance values (yellow/white zones) strongly correlate with rising temperatures, adhering to the fundamental principle of Solar Heating. Conversely, the density of points at low irradiance/temperature corresponds accurately to morning/evening transitions. This confirms that the network has successfully learned the thermodynamic relationship between heat and light.

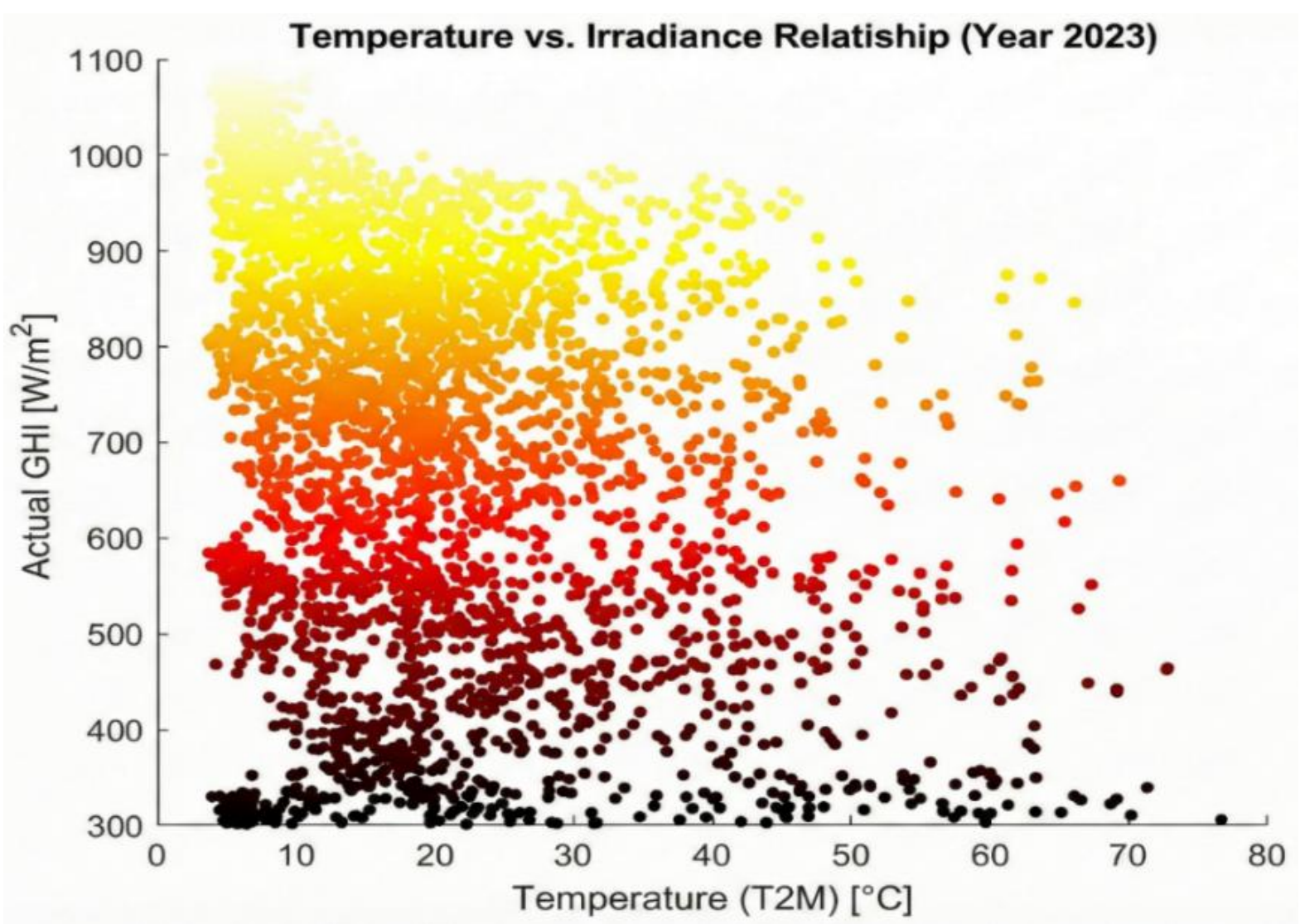

**Fig. 17.** Demonstration of physical consistency, where the predictions adhered to the theoretical Clear Sky Limit.

2. Boundary Constraints:

Furthermore, Fig. 18 demonstrates that the model strictly respects the Clear-Sky Limit. The predictions never exceed the theoretical maximum radiation curve, proving that the physics-guided features effectively act as "soft constraints" that bound the neural network's output within physically possible limits.

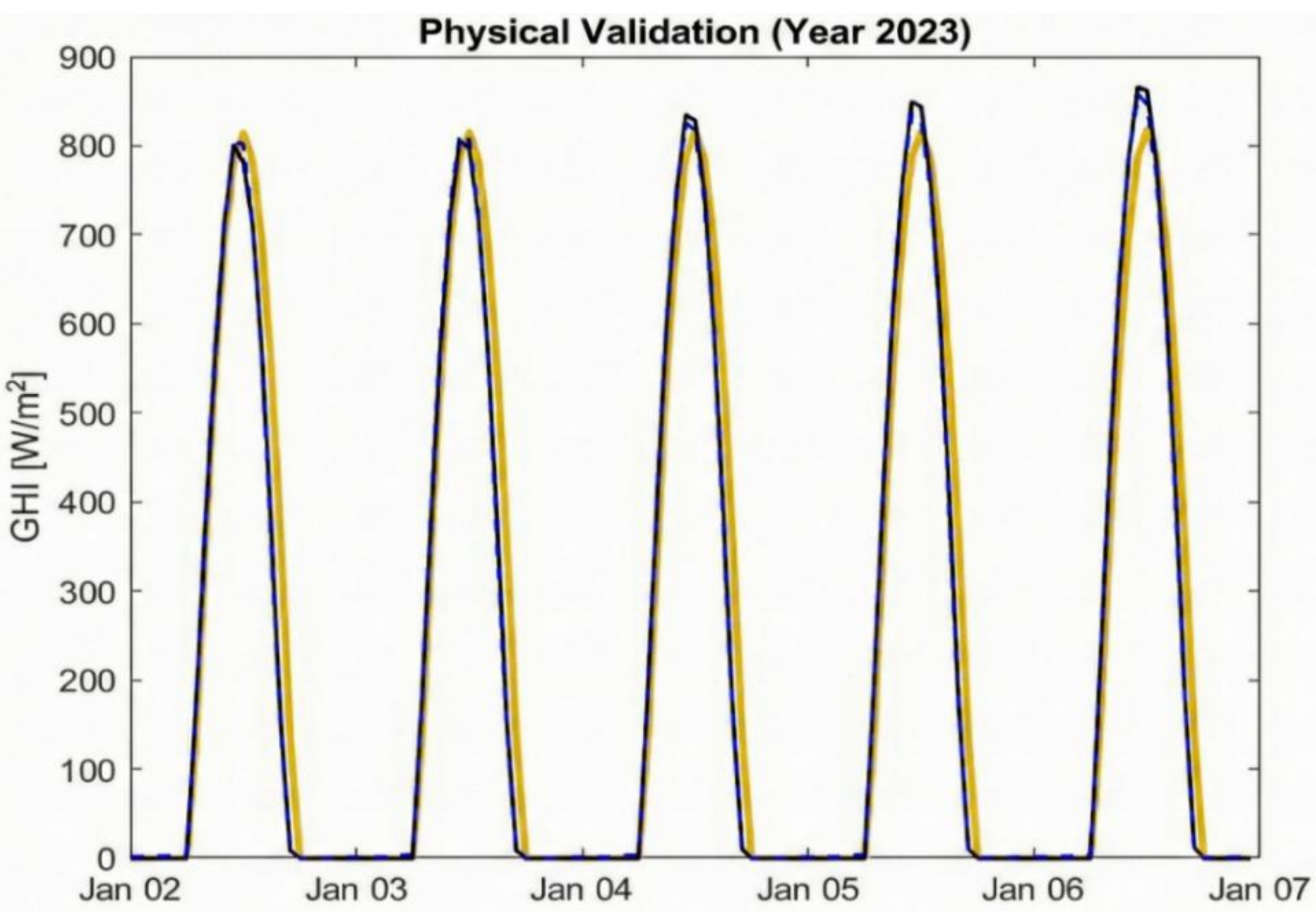

**Fig. 18.** Validation of the physical consistency of the predicted GHI values.

### 3.7 Ablation Study: Empirical Verification of the Complexity Paradox

To empirically verify the limitations of structural complexity discussed in Section 3.5, we conducted a controlled experiment by deliberately augmenting our Physics-Informed model with a Self-Attention Mechanism. The objective was to test whether adding computational depth yields better accuracy than physical guidance.

1. Experimental Setup:

We compared the optimal Proposed PI-Hybrid (CNN-BiLSTM + Physics) against an Attention-Augmented Variant (CNN-BiLSTM - Attention + Physics). Both models were trained on the same dataset with identical hyperparameters to ensure a fair comparison.

2. Quantitative Results:

As summarized in Table 7, the results confirm the "Complexity Paradox."

- Performance Degradation: The addition of the Self-Attention layer increased the RMSE from 19.53 W/m² to 30.64 W/m², and reduced the $R^2$ score from 0.997 to 0.992.
- Error Analysis: As illustrated in the error distribution plots (Fig. 19), the attention-based variant exhibited significant instability during peak irradiance hours.

3. Conclusion:

This experiment provides the concrete mathematical proof for our hypothesis: when explicit physical precursors (like Clear-Sky Index) are provided, the "weight focusing" task is already solved. Consequently, the Self-Attention mechanism becomes redundant, introducing computational noise rather than extracting new features. This validates that Explicit Physics is a superior, deterministic "Attention Mechanism" for solar forecasting in arid environments.

**Table 7.** Summary of models in the ablation and complexity study

| Model Architecture | Physics-Informed | Attention Mechanism | RMSE (W/m²) | $R^2$ |
|---|---|---|---|---|
| Standard Baseline | No | No | 55.32 | 0.970 |
| Attention-Hybrid | Yes | Yes | 30.64 | 0.992 |
| Proposed PI-Hybrid | Yes | No | 19.53 | 0.997 |

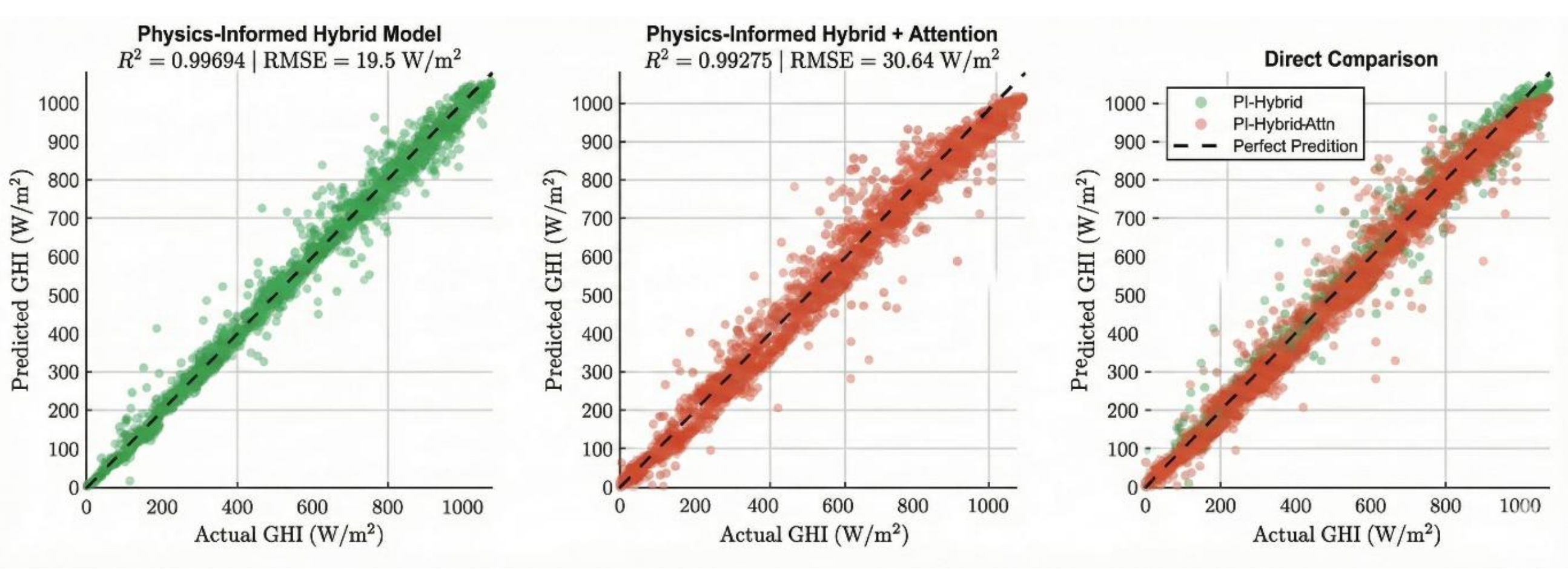

**Fig. 19.** Performance comparison between the proposed model and the attention-based model, illustrating the accuracy degradation with increased complexity.

Furthermore, the temporal error analysis in Fig. 20 highlights that the attention-based model suffers from significant deviations during peak noon hours and summer months. In contrast, the proposed physics-guided framework maintains high stability during these periods, proving that deterministic physical constraints act as a superior "attention" mechanism for solar irradiance tasks.

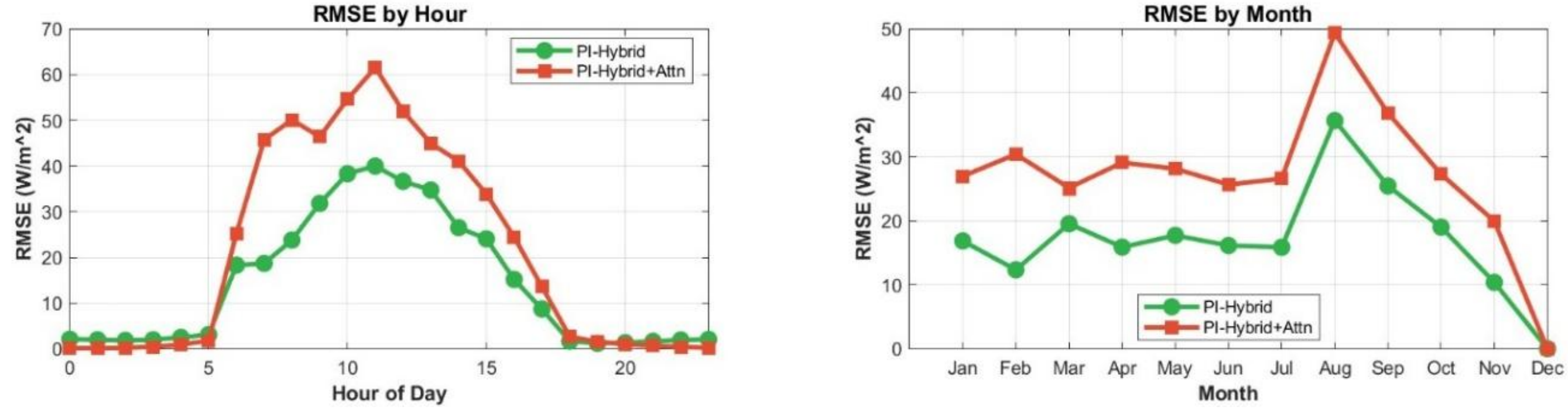

**Fig. 20.** Temporal error analysis, highlighting the superiority of the proposed model during peak periods.

### 3.8 Long-Term Robustness and Generalization (2020–2024)

To ensure the operational validity of the framework in a changing climate, a long-term Stress Test was conducted on an independent dataset spanning five recent years (2020–2024). The annual results, summarized in Fig. 21, demonstrate exceptional stability; the coefficient of determination ($R^2$) remained consistently above 0.995 throughout the entire five-year horizon.

### 3.9 Summary of Forecasting Results

The comprehensive analysis confirms that the proposed PI-Hybrid framework represents the optimal solution, striking a balance between high predictive fidelity and structural simplicity. The physics-guided approach successfully eradicated the common pitfalls of traditional data-driven models, such as phase lag and poor tracking during cloudy days. Furthermore, the model demonstrated exceptional temporal stability over long horizons, establishing it as a reliable and robust foundation for the predictive control system presented in the subsequent section.

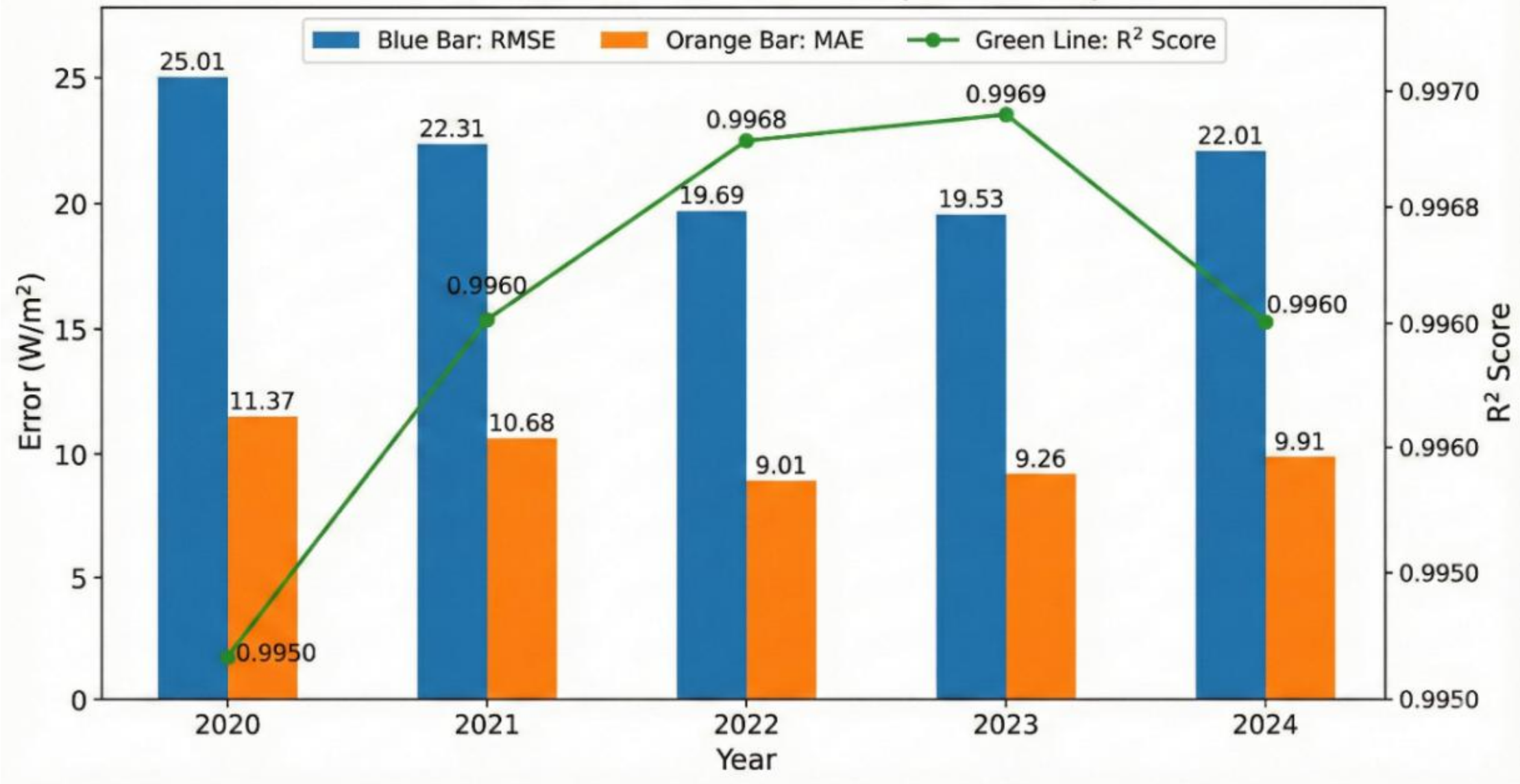

**Fig. 21.** Evolution of model accuracy across the testing years (2020-2024).

## 4. Conclusions

Based on the extensive training and validation conducted within the MATLAB/Simulink environment, core conclusions were drawn regarding the feasibility and architectural efficiency of the proposed system. The study attributes the superior predictive power of the model to the strategic selection of 15 spatiotemporal variables, concluding that this feature set is practical and economically viable. Zero-cost algorithmic features, comprising nearly half of the inputs such as physics-derived indicators and cyclical time encodings, act as the physical compass of the model. These features embed the deterministic laws of solar geometry and temporal periodicity directly into the processor without requiring external sensors. Primary variables such as global horizontal irradiance and ambient temperature provide immediate correlation with the local environment and can be measured locally using low-cost industrial sensors. Furthermore, complex variables like the satellite clearness index, direct normal irradiance, and diffuse horizontal irradiance were found critical for recognizing cloud signatures and dust effects, and can be fetched via application programming interfaces (APIs) to offer a cost-effective pathway for enhanced energy planning.

The results justified the use of a deep hybrid architecture combining convolutional neural networks and bidirectional long short-term memory (CNN-BiLSTM) as essential for mapping high-volatility features. However, a significant finding emerged regarding structural complexity. The ablation study revealed that the addition of a self-attention layer placed after the BiLSTM layer resulted in a degradation in accuracy, increasing the root mean square error (RMSE) from 19.5 to 30.64 W/m², rather than an improvement. This finding demonstrates that pre-injected physical guidance intrinsically performs the task of weight focusing on critical moments more effectively than learned weights. This empirically validates the hypothesis that explicit physics is a superior attention mechanism for solar forecasting, offering a dual advantage of higher accuracy and significantly reduced computational overhead compared to heavier transformer-based alternatives (Al-Ali et al., 2023).

While the proposed physics-informed framework serves as a high-precision teacher model, its computational footprint remains a challenge for resource-constrained controllers. Future work will focus on transitioning this intelligence to the edge to enable real-time, autonomous control of solar pumping stations through a three-tier compression strategy. This includes knowledge distillation to train a lightweight student network to mimic the feature-extraction capabilities of the CNN-BiLSTM model with lower latency (Hinton et al., 2015). Additionally, structured pruning and quantization leveraging the lottery ticket hypothesis will be used to identify and remove redundant synaptic connections within the BiLSTM layers (Frankle and Carbin, 2019). Finally, applying numerical integer quantization will reduce the memory footprint of the model, allowing it to run on low-cost microcontrollers to protect motor windings from sudden irradiance-induced ramp-rate events.

A transition from reactive control to an integrated proactive irrigation framework is recommended. This strategy leverages the 5-hour predictive horizon and 24-hour historical window of the model to implement a multi-modal storage approach. During peak irradiance, the system executes pre-emptive pumping to convert excess solar energy into virtual storage within the soil matrix and water reservoirs, reducing reliance on costly chemical storage. To ensure long-term resilience, a hierarchical AI-driven control is suggested for managing strategic water-soil balances, while a low-level model predictive controller ensures real-time stabilization and soft-braking protection for the variable frequency drive. Future iterations should shift toward producing confidence intervals rather than deterministic point forecasts, providing an uncertainty score that allows irrigation controllers to make risk-aware decisions, such as delaying sensitive fertigation processes during periods of high atmospheric volatility to prevent hardware clogging and ensure uniform nutrient distribution.

## Acknowledgements

The authors would like to express sincere gratitude to Eng. Wail Motwakil Idrees for his valuable guidance and constructive advice during the development of this project. His technical insights regarding neural network optimization and hybrid forecasting strategies were instrumental in refining the methodology and improving the overall quality of this work.